\documentstyle[aclap]{article}

\newcommand{\SAC}{{\small \sf SAC}}
\newcommand{\SIC}{{\small \sf SIC}}
\newcommand{\DTG}{{\small \sf DTG}}
\newcommand{\XTAG}{{\small \sf XTAG}}
\newcommand{\SAtree}{{\small \sf SA}-tree}
\newcommand{\CKY}{{\small \sf CKY}}
\newcommand{\CFG}{{\small \sf CFG}}
\newcommand{\UVGDL}{{\small \sf UVG-DL}}
\newcommand{\HPSG}{{\small \sf HPSG}}
\newcommand{\MCTAGDL}{{\small \sf MCTAG-DL}}
\newcommand{\MCTAG}{{\small \sf MCTAG}}
\newcommand{\VTAG}{{\small \sf V-TAG}}
\newcommand{\IDLP}{{\small \sf ID/LP}}
\newcommand{\LFG}{{\small \sf LFG}}
\newcommand{\TAG}{{\small \sf TAG}}
\newcommand{\LTAG}{{\small \sf LTAG}}
\newcommand{\ENA}{{\small \sf ENA}}

\newcommand	{\set}[1]	{\left\{\,{#1}\,\right\}}

\newcommand{\featf}{\sf}
\makeatletter

\@ifundefined{LaTeXe}{}{
   \let\normalsize\@normalsize
}

\newcounter{enums}
\def\enumsentence{\@ifnextchar[{\@enumsentence}
{\refstepcounter{enums}\@enumsentence[(\theenums)]}}
\long\def\@enumsentence[#1]#2{\begin{list}{}{\topsep=1ex%
					     \itemsep=1ex}
\item[#1] {#2}
\end{list}}
\newcounter{tempcnt}

\def\@item[#1]{\if@noparitem \@donoparitem
  \else \if@inlabel \indent \par \fi
         \ifhmode \unskip\unskip \par \fi
         \if@newlist \if@nobreak \@nbitem \else
                        \addpenalty\@beginparpenalty
                        \addvspace\@topsep \addvspace{-\parskip}\fi
           \else \addpenalty\@itempenalty \addvspace\itemsep
          \fi
    \global\@inlabeltrue
\fi
\everypar{\global\@minipagefalse\global\@newlistfalse
          \if@inlabel\global\@inlabelfalse \hskip -\parindent \box\@labels
             \penalty\z@ \fi
          \everypar{}}\global\@nobreakfalse
\if@noitemarg \@noitemargfalse \if@nmbrlist \refstepcounter{\@listctr}\fi \fi
\setbox\@tempboxa\hbox{\makelabel{#1}}%
\global\setbox\@labels
 \hbox{\unhbox\@labels \hskip \itemindent
       \hskip -\labelwidth \hskip -\labelsep
       \ifdim \wd\@tempboxa >\labelwidth
                \box\@tempboxa
          \else \hbox to\labelwidth {\unhbox\@tempboxa}\fi
       \hskip \labelsep}\ignorespaces}

\newcounter{enumsi}

\def\@mklab#1{\hfil#1}
\def\enummklab#1{\hfil(\eelabel)\hbox to 15pt{\hfil#1}}
\def\enummakelabel#1{\enummklab{#1}\global\let\makelabel=\@mklab}

\def\eenumsentence{\@ifnextchar[{\@eenumsentence}
{\refstepcounter{enums}\@eenumsentence[\theenums]}}

\long\def\@eenumsentence[#1]#2{\def\eelabel{#1}%
\begin{list}{\alph{enumsi}.}{\usecounter{enumsi}%
\advance\leftmargin by 15pt\advance\labelwidth by 15pt%
\let\makelabel=\enummakelabel%
\topsep=1ex%
\itemsep=1ex}
{#2}
\end{list}}

\newcommand{\si}{\mbox{$_i$}}

\newcommand{\ti}{t$_i$}

\newcommand{\nom}{\mbox{$_{\mbox{\sc \tiny NOM}}$\hspace{0.8em}}}
\newcommand{\erg}{\mbox{$_{\mbox{\sc \tiny ERG}}$\hspace{0.8em}}}
\newcommand{\dat}{\mbox{$_{\mbox{\sc \tiny DAT}}$\hspace{0.8em}}}

\newcommand{\nperf}{\mbox{$_{\mbox{\sc \tiny NPerf}}$\hspace{0.8em}}}

\newcommand{\es}[1]{\enumsentence{ #1}}
\newcommand{\ees}[1]{\eenumsentence{ #1}}

\let\@gsingle=1
\def\singlegloss{\let\@gsingle=1}
\def\nosinglegloss{\let\@gsingle=0}
\@ifundefined{new@fontshape}%
   {\def\@selfnt{\ifx\@currsize\normalsize\@normalsize\else\@currsize\fi}}
   {\def\@selfnt{\selectfont}}

\def\gll
   {\begin{flushleft}
     \ifx\@gsingle1
        \vskip\baselineskip\def\baselinestretch{1}%
        \@selfnt\vskip-\baselineskip\fi%
    \bgroup
    \twosent
   }

\def\glll
   {\begin{flushleft}
     \ifx\@gsingle1
        \vskip\baselineskip\def\baselinestretch{1}%
        \@selfnt\vskip-\baselineskip\fi%
    \bgroup
    \threesent
   }

\def\glt{\vskip.17\baselineskip}

\newbox\lineone
\newbox\linetwo%
\newbox\linethree%
\newbox\wordone
\newbox\wordtwo%
\newbox\wordthree%
\newbox\gline
\newskip\glossglue
\newif\ifnotdone

\@ifundefined{eachwordone}{\let\eachwordone=\rm}{\relax}
\@ifundefined{eachwordtwo}{\let\eachwordtwo=\rm}{\relax}
\@ifundefined{eachwordthree}{\let\eachwordthree=\rm}{\relax}

\def\lastword#1#2#3
   {\setbox#2=\vbox{\unvbox#2%
                    \global\setbox#3=\lastbox%
                   }%
    \ifvoid#3\global\setbox#3=\hbox{#1\strut{} }\fi
   }

\def\testdone
   {\ifdim\ht\lineone=0pt
         \ifdim\ht\linetwo=0pt \notdonefalse 
         \else\notdonetrue
         \fi
    \else\notdonetrue
    \fi
   }

\gdef\getwords(#1,#2)#3 #4\\
   {\setbox#1=\vbox{\hbox{#2\strut#3 }
                    \unvbox#1%
                   }%
    \def\more{#4}%
    \ifx\more\empty\let\more=\donewords
    \else\let\more=\getwords
    \fi
    \more(#1,#2)#4\\%
   }

\gdef\donewords(#1,#2)\\{}%

\gdef\twosent#1\\ #2\\{
    \getwords(\lineone,\eachwordone)#1 \\%
    \getwords(\linetwo,\eachwordtwo)#2 \\%
    \loop\lastword{\eachwordone}{\lineone}{\wordone}%
         \lastword{\eachwordtwo}{\linetwo}{\wordtwo}%
         \global\setbox\gline=\hbox{\unhbox\gline
                                    \hskip\glossglue
                                    \vtop{\box\wordone   
                                          \nointerlineskip
                                          \box\wordtwo
                                         }%
                                   }%
         \testdone
         \ifnotdone
    \repeat
    \egroup 
   \gl@stop}

\gdef\threesent#1\\ #2\\ #3\\{
    \getwords(\lineone,\eachwordone)#1 \\%
    \getwords(\linetwo,\eachwordtwo)#2 \\%
    \getwords(\linethree,\eachwordthree)#3 \\%
    \loop\lastword{\eachwordone}{\lineone}{\wordone}%
         \lastword{\eachwordtwo}{\linetwo}{\wordtwo}%
         \lastword{\eachwordthree}{\linethree}{\wordthree}%
         \global\setbox\gline=\hbox{\unhbox\gline
                                    \hskip\glossglue
                                    \vtop{\box\wordone   
                                          \nointerlineskip
                                          \box\wordtwo
                                          \nointerlineskip
                                          \box\wordthree
                                         }%
                                   }%
         \testdone
         \ifnotdone
    \repeat
    \egroup 
   \gl@stop}

\def\gl@stop{{\hskip -\glossglue}\unhbox\gline\end{flushleft}}

\makeatother

\title{D-Tree Grammars}

\author{
Owen Rambow\\
CoGenTex, Inc.\\
840 Hanshaw Road\\
Ithaca, NY 14850\\
{\tt owen@cogentex.com}
\And
K. Vijay-Shanker\\
Department of Computer \&\\
Information Science\\
University of Delaware\\
Newark, DE 19716\\
{\tt vijay@udel.edu}
\And
David Weir\\
School of Cognitive \& \\
Computing Sciences\\
University of Sussex\\
Brighton, BN1 9HQ, UK.\\
{\tt david.weir@cogs.susx.ac.uk}
}

\begin{document}

\maketitle
\bibliographystyle{acl}

\begin{abstract}

{\DTG} are designed to share some of the advantages of {\TAG}
while overcoming some of its limitations.  {\DTG} involve two
composition operations called subsertion and sister-adjunction. The most
distinctive feature of {\DTG} is that, unlike {\TAG}, there is complete
uniformity in the way that the two {\DTG} operations relate lexical
items: subsertion always corresponds to complementation and
sister-adjunction to modification.  Furthermore, {\DTG}, unlike {\TAG},
can provide a uniform analysis for {\em wh}-movement in English and
Kashmiri, despite the fact that the {\em wh} element in Kashmiri appears
in sentence-second position, and not sentence-initial position as in
English.

\end{abstract}

\section{Introduction}
\label{sec-intro}

We define a new grammar formalism, called D-Tree Grammars~({\DTG}), which
arises from work on Tree-Adjoining Grammars~({\TAG})~\cite{jlt75}.  A
salient feature of {\TAG} is the extended domain of locality it
provides.  Each elementary structure can be associated with a lexical
item (as in Lexicalized {\TAG}~({\LTAG})~\cite{joshi/schabes:1991}).
Properties related to the lexical item (such as subcategorization,
agreement, certain types of word order variation) can be expressed
within the elementary structure~\cite{k87,frank:1992}.  In addition, {\TAG}
remain tractable, yet their generative capacity is sufficient to account
for certain syntactic phenomena that, it has been argued, lie beyond
Context-Free Grammars~({\CFG})~\cite{sh85b}.  {\TAG}, however, has two
limitations which provide the motivation for this work. The first
problem (discussed in Section~\ref{sec-der-dep}) is that the {\TAG}
operations of substitution and adjunction do not map cleanly onto the
relations of complementation and modification. A second problem
(discussed in Section~\ref{sec-prob-cons}) has to do with the inability
of {\TAG} to provide analyses for certain syntactic phenomena.  In
developing {\DTG} we have tried to overcome these problems while remaining
faithful to what we see as the key advantages of {\TAG} (in particular, its
enlarged domain of locality). In Section~\ref{sec-dtg} we introduce some
of the key features of {\DTG} and explain how they are intended to address
the problems that we have identified with {\TAG}\@.

\subsection{Derivations and Dependencies}
\label{sec-der-dep}

In {\LTAG}, the operations of substitution and adjunction relate two
lexical items.  It is therefore natural to interpret these operations as
establishing a direct linguistic relation between the two lexical items,
namely a relation of complementation (predicate-argument relation) or of
modification.  In purely {\CFG}-based approaches, these relations are
only implicit.  However, they represent important linguistic intuition,
they provide a uniform interface to semantics, and they are, as Schabes
\& Shieber~\shortcite{ss94} argue, important in order to support
statistical parameters in stochastic frameworks and appropriate
adjunction constraints in {\TAG}\@.  In many frameworks, complementation
and modification are in fact made explicit:
{\LFG}~\cite{bresnan/kaplan:1982} provides a separate functional (f-)
structure, and dependency grammars (see e.g. Mel'{\v c}uk (1988)) use
these notions as the principal basis for syntactic representation.  We
will follow the dependency literature in referring to complementation
and modification as syntactic dependency.  As observed by Rambow and
Joshi~\shortcite{rambow/joshi:1992}, for {\TAG}, the importance of the
dependency structure means that not only the derived phrase-structure
tree is of interest, but also the operations by
which we obtained it from elementary structures.
This information is encoded in the
derivation tree~\cite{v87}.

However, as Vijay-Shanker~\shortcite{vijay92} observes, the {\TAG}
composition operations are not used uniformly: while substitution is
used only to add a (nominal) complement, adjunction is used both for
modification and (clausal) complementation. Clausal complementation
could not be handled uniformly by substitution because of the existence
of syntactic phenomena such as long-distance {\em wh}-movement in
English.  Furthermore, there is an inconsistency in the directionality
of the operations used for complementation in {\TAG}@: nominal
complements are substituted into their governing verb's tree, while the
governing verb's tree is adjoined into its own clausal complement.  The
fact that adjunction and substitution are used in a linguistically
heterogeneous manner means that (standard) {\TAG} derivation trees do
not provide a good representation of the dependencies between the words
of the sentence, i.e., of the predicate-argument and modification
structure.

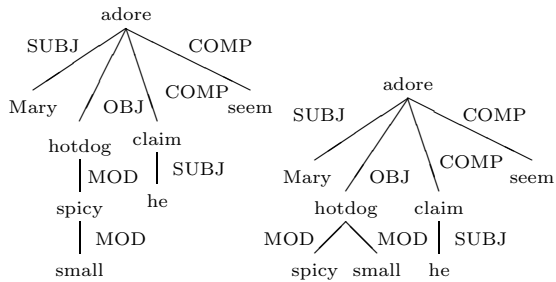
\begin{figure}[htb]
\begin{center}
\setlength{\unitlength}{0.008125in}%
\begingroup\makeatletter\ifx\SetFigFont\undefined
\def\x#1#2#3#4#5#6#7\relax{\def\x{#1#2#3#4#5#6}}%
\expandafter\x\fmtname xxxxxx\relax \def\y{splain}%
\ifx\x\y   
\gdef\SetFigFont#1#2#3{%
  \ifnum #1<17\tiny\else \ifnum #1<20\small\else
  \ifnum #1<24\normalsize\else \ifnum #1<29\large\else
  \ifnum #1<34\Large\else \ifnum #1<41\LARGE\else
     \huge\fi\fi\fi\fi\fi\fi
  \csname #3\endcsname}%
\else
\gdef\SetFigFont#1#2#3{\begingroup
  \count@#1\relax \ifnum 25<\count@\count@25\fi
  \def\x{\endgroup\@setsize\SetFigFont{#2pt}}%
  \expandafter\x
    \csname \romannumeral\the\count@ pt\expandafter\endcsname
    \csname @\romannumeral\the\count@ pt\endcsname
  \csname #3\endcsname}%
\fi
\fi\endgroup
\begin{picture}(164,178)(248,600)
\thinlines
\put(290,675){\line( 0,-1){ 20}}
\put(290,635){\line( 0,-1){ 20}}
\put(290,680){\makebox(0,0)[b]{\smash{\SetFigFont{7}{8.4}{rm}hotdog}}}
\put(295,660){\makebox(0,0)[lb]{\smash{\SetFigFont{7}{8.4}{rm}MOD}}}
\put(290,640){\makebox(0,0)[b]{\smash{\SetFigFont{7}{8.4}{rm}spicy}}}
\put(290,600){\makebox(0,0)[b]{\smash{\SetFigFont{7}{8.4}{rm}small}}}
\put(300,620){\makebox(0,0)[lb]{\smash{\SetFigFont{7}{8.4}{rm}MOD}}}
\put(320,760){\line( 1,-3){ 20}}
\put(320,760){\line( 2,-1){ 80}}
\put(320,760){\line(-3,-2){ 60}}
\put(320,760){\line(-1,-2){ 30}}
\put(340,680){\line( 0,-1){ 20}}
\put(320,765){\makebox(0,0)[b]{\smash{\SetFigFont{7}{8.4}{rm}adore}}}
\put(360,745){\makebox(0,0)[lb]{\smash{\SetFigFont{7}{8.4}{rm}COMP}}}
\put(400,705){\makebox(0,0)[b]{\smash{\SetFigFont{7}{8.4}{rm}seem}}}
\put(340,685){\makebox(0,0)[b]{\smash{\SetFigFont{7}{8.4}{rm}claim}}}
\put(340,645){\makebox(0,0)[b]{\smash{\SetFigFont{7}{8.4}{rm}he}}}
\put(305,705){\makebox(0,0)[lb]{\smash{\SetFigFont{7}{8.4}{rm}OBJ}}}
\put(260,705){\makebox(0,0)[b]{\smash{\SetFigFont{7}{8.4}{rm}Mary}}}
\put(290,745){\makebox(0,0)[rb]{\smash{\SetFigFont{7}{8.4}{rm}SUBJ}}}
\put(350,665){\makebox(0,0)[lb]{\smash{\SetFigFont{7}{8.4}{rm}SUBJ}}}
\put(345,715){\makebox(0,0)[lb]{\smash{\SetFigFont{7}{8.4}{rm}COMP}}}
\end{picture}
\setlength{\unitlength}{0.008125in}%
\begingroup\makeatletter\ifx\SetFigFont\undefined
\def\x#1#2#3#4#5#6#7\relax{\def\x{#1#2#3#4#5#6}}%
\expandafter\x\fmtname xxxxxx\relax \def\y{splain}%
\ifx\x\y   
\gdef\SetFigFont#1#2#3{%
  \ifnum #1<17\tiny\else \ifnum #1<20\small\else
  \ifnum #1<24\normalsize\else \ifnum #1<29\large\else
  \ifnum #1<34\Large\else \ifnum #1<41\LARGE\else
     \huge\fi\fi\fi\fi\fi\fi
  \csname #3\endcsname}%
\else
\gdef\SetFigFont#1#2#3{\begingroup
  \count@#1\relax \ifnum 25<\count@\count@25\fi
  \def\x{\endgroup\@setsize\SetFigFont{#2pt}}%
  \expandafter\x
    \csname \romannumeral\the\count@ pt\expandafter\endcsname
    \csname @\romannumeral\the\count@ pt\endcsname
  \csname #3\endcsname}%
\fi
\fi\endgroup
\begin{picture}(176,133)(236,645)
\thinlines
\put(280,680){\line(-1,-1){ 20}}
\put(280,680){\line( 1,-1){ 20}}
\put(280,685){\makebox(0,0)[b]{\smash{\SetFigFont{7}{8.4}{rm}hotdog}}}
\put(260,665){\makebox(0,0)[rb]{\smash{\SetFigFont{7}{8.4}{rm}MOD}}}
\put(300,665){\makebox(0,0)[lb]{\smash{\SetFigFont{7}{8.4}{rm}MOD}}}
\put(260,645){\makebox(0,0)[b]{\smash{\SetFigFont{7}{8.4}{rm}spicy}}}
\put(300,645){\makebox(0,0)[b]{\smash{\SetFigFont{7}{8.4}{rm}small}}}
\put(320,760){\line( 2,-1){ 80}}
\put(320,760){\line( 1,-3){ 20}}
\put(320,760){\line(-3,-2){ 60}}
\put(320,760){\line(-2,-3){ 40}}
\put(340,680){\line( 0,-1){ 20}}
\put(320,765){\makebox(0,0)[b]{\smash{\SetFigFont{7}{8.4}{rm}adore}}}
\put(400,705){\makebox(0,0)[b]{\smash{\SetFigFont{7}{8.4}{rm}seem}}}
\put(355,745){\makebox(0,0)[lb]{\smash{\SetFigFont{7}{8.4}{rm}COMP}}}
\put(255,705){\makebox(0,0)[b]{\smash{\SetFigFont{7}{8.4}{rm}Mary}}}
\put(295,705){\makebox(0,0)[lb]{\smash{\SetFigFont{7}{8.4}{rm}OBJ}}}
\put(340,685){\makebox(0,0)[b]{\smash{\SetFigFont{7}{8.4}{rm}claim}}}
\put(350,665){\makebox(0,0)[lb]{\smash{\SetFigFont{7}{8.4}{rm}SUBJ}}}
\put(340,645){\makebox(0,0)[b]{\smash{\SetFigFont{7}{8.4}{rm}he}}}
\put(340,715){\makebox(0,0)[lb]{\smash{\SetFigFont{7}{8.4}{rm}COMP}}}
\put(280,745){\makebox(0,0)[rb]{\smash{\SetFigFont{7}{8.4}{rm}SUBJ}}}
\end{picture}
\end{center}
\caption{Derivation trees for (\protect\ref{exa-english}): original
definition (left); Schabes \& Shieber definition (right)}
\label{fig-dep-bad}
\end{figure}

\begin{figure}[hbt]
\begin{center}
\setlength{\unitlength}{0.008125in}%
\begingroup\makeatletter\ifx\SetFigFont\undefined
\def\x#1#2#3#4#5#6#7\relax{\def\x{#1#2#3#4#5#6}}%
\expandafter\x\fmtname xxxxxx\relax \def\y{splain}%
\ifx\x\y   
\gdef\SetFigFont#1#2#3{%
  \ifnum #1<17\tiny\else \ifnum #1<20\small\else
  \ifnum #1<24\normalsize\else \ifnum #1<29\large\else
  \ifnum #1<34\Large\else \ifnum #1<41\LARGE\else
     \huge\fi\fi\fi\fi\fi\fi
  \csname #3\endcsname}%
\else
\gdef\SetFigFont#1#2#3{\begingroup
  \count@#1\relax \ifnum 25<\count@\count@25\fi
  \def\x{\endgroup\@setsize\SetFigFont{#2pt}}%
  \expandafter\x
    \csname \romannumeral\the\count@ pt\expandafter\endcsname
    \csname @\romannumeral\the\count@ pt\endcsname
  \csname #3\endcsname}%
\fi
\fi\endgroup
\begin{picture}(185,173)(147,525)
\thinlines
\put(200,680){\line(-2,-1){ 40}}
\put(200,680){\line( 2,-1){ 40}}
\put(240,640){\line( 0,-1){ 20}}
\put(240,600){\line(-2,-1){ 40}}
\put(240,600){\line( 2,-1){ 40}}
\put(280,560){\line(-2,-1){ 40}}
\put(280,560){\line( 2,-1){ 40}}
\put(200,685){\makebox(0,0)[b]{\smash{\SetFigFont{7}{8.4}{rm}claim}}}
\put(160,645){\makebox(0,0)[b]{\smash{\SetFigFont{7}{8.4}{rm}he}}}
\put(230,670){\makebox(0,0)[lb]{\smash{\SetFigFont{7}{8.4}{rm}COMP}}}
\put(170,670){\makebox(0,0)[rb]{\smash{\SetFigFont{7}{8.4}{rm}SUBJ}}}
\put(240,645){\makebox(0,0)[b]{\smash{\SetFigFont{7}{8.4}{rm}seem}}}
\put(240,605){\makebox(0,0)[b]{\smash{\SetFigFont{7}{8.4}{rm}adore}}}
\put(250,625){\makebox(0,0)[lb]{\smash{\SetFigFont{7}{8.4}{rm}COMP}}}
\put(200,565){\makebox(0,0)[b]{\smash{\SetFigFont{7}{8.4}{rm}Mary}}}
\put(205,590){\makebox(0,0)[rb]{\smash{\SetFigFont{7}{8.4}{rm}SUBJ}}}
\put(270,590){\makebox(0,0)[lb]{\smash{\SetFigFont{7}{8.4}{rm}OBJ}}}
\put(280,565){\makebox(0,0)[b]{\smash{\SetFigFont{7}{8.4}{rm}hotdog}}}
\put(250,550){\makebox(0,0)[rb]{\smash{\SetFigFont{7}{8.4}{rm}MOD}}}
\put(305,550){\makebox(0,0)[lb]{\smash{\SetFigFont{7}{8.4}{rm}MOD}}}
\put(320,525){\makebox(0,0)[b]{\smash{\SetFigFont{7}{8.4}{rm}small}}}
\put(240,525){\makebox(0,0)[b]{\smash{\SetFigFont{7}{8.4}{rm}spicy}}}
\end{picture}
\end{center}
\caption{Dependency tree for (\protect\ref{exa-english})}
\label{fig-dep}
\end{figure}
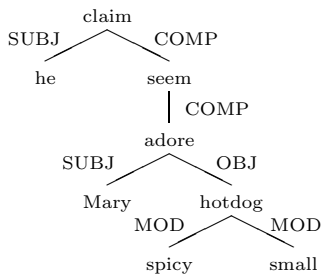

For instance, English sentence (\ref{exa-english}) gets the derivation
structure shown on the left in
Figure~\ref{fig-dep-bad}\protect\footnote{For clarity, we depart from
standard {\TAG} notational practice and annotate nodes with lexemes and
arcs with grammatical function.}.
{\small \es{\label{exa-english}Small spicy hotdogs he claims Mary seems to
adore} }
When comparing this derivation structure to the dependency structure in
Figure~\ref{fig-dep}, the following problems become apparent. First,
both adjectives depend on {\em hotdog}, while in the derivation
structure {\em small} is a daughter of {\em spicy}. In addition,
{\em seem} depends on {\em claim} (as does its nominal argument, {\em
he}), and {\em adore} depends on {\em seem}. In the derivation
structure, {\em seem} is a daughter of {\em adore} (the direction does
not express the actual dependency), and {\em claim} is also a daughter
of {\em adore} (though neither is an argument of the other).

Schabes \& Shieber~\shortcite{ss94} solve the first problem by
distinguishing between the adjunction of modifiers and of clausal
complements.  This gives us the derivation structure shown on the right
in Figure~\ref{fig-dep-bad}.  While this might provide a satisfactory
treatment of modification at the derivation level, there are now three
types of operations (two adjunctions and substitution) for two types of
dependencies (arguments and modifiers), and the directionality problem
for embedded clauses remains unsolved.

In defining {\DTG} we have attempted to resolve these problems with the use
of a single operation (that we call subsertion) for handling all
complementation and a second operation (called sister-adjunction)
for modification.  Before discussion these operations further we
consider a second problem with {\TAG} that has implications for the design
of these new composition operations (in particular, subsertion).

\subsection{Problematic Constructions for {\TAG}}
\label{sec-prob-cons}

{\TAG} cannot be used to provide suitable analyses for certain syntactic
phenomena, including long-distance scrambling in German~\cite{bjr91},
Romance Clitics~\cite{bleam94}, {\em wh}-extraction out of
complex picture-NPs~\cite{k87}, and Kashmiri {\em wh}-extraction (presented
here).  The problem in describing these phenomena with
{\TAG} arises from the fact (observed by Vijay-Shanker~\shortcite{vijay92})
that adjoining is an overly restricted way of combining structures.  We
illustrate the problem by considering Kashmiri {\em wh}-extraction,
drawing on Bhatt~\shortcite{bhatt:1994}.  {\em Wh}-extraction in
Kashmiri proceeds as in English, except that the {\em wh}-word ends up
in sentence-second position, with a topic from the matrix clause in
sentence-initial position.  This is illustrated in (\ref{sent-kash}a)
for a simple clause and in (\ref{sent-kash}b) for a complex clause.

{\small
\ees{\label{sent-kash}
\item
\gll
rameshan kyaa dyutnay tse\\                     
{Ramesh\erg} {what\nom} gave {you\dat}\\
\glt
What did you give Ramesh?
\item
\gll						
rameshan {kyaa\si} chu baasaan {[} ki me kor {\ti]}\\
{Ramesh\erg} what is {believe\nperf} {} that {I\erg} do {}\\
\glt
What does Ramesh believe that I did?
}
}

Since the moved element does not appear in sentence-initial position,
the {\TAG} analysis of English {\em wh}-extraction of
Kroch~\shortcite{k87,kroch:89a} (in which the matrix clause is adjoined
into the embedded clause) cannot be transferred, and in fact no
linguistically plausible {\TAG} analysis appears to be available.

In the past, variants of {\TAG} have been developed to extend the range of
possible analyses.  In Multi-Component {\TAG}~({\MCTAG})~\cite{j87b}, trees
are grouped into sets which must be adjoined together (multicomponent
adjunction). However, {\MCTAG} lack expressive power since, while syntactic
relations are invariably subject to c-command or dominance constraints,
there is no way to state that two trees from a set must be in a
dominance relation in the derived tree.  {\MCTAG} with Domination
Links~({\MCTAGDL})~\cite{bjr91} are multicomponent systems that allow for
the expression of dominance constraints.  However, {\MCTAGDL} share a
further problem with {\MCTAG}: the derivation structures cannot be given a
linguistically meaningful interpretation. Thus, they fail to address the
first problem we discussed (in Section~\ref{sec-der-dep}).

\subsection{The {\DTG} Approach}
\label{sec-dtg}

Vijay-Shanker~\shortcite{vijay92} points out that use of adjunction for
clausal complementation in {\TAG} corresponds, at the level of dependency
structure, to substitution at the foot node\footnote{In these cases the
foot node is an argument node of the lexical anchor.} of the adjoined tree.
However, adjunction (rather than substitution) is used since, in general,
the structure that is substituted may only form part of the clausal
complement: the remaining substructure of the clausal complement appears
above the root of the adjoined tree.  Unfortunately, as seen in the
examples given in Section~\ref{sec-prob-cons}, there are cases where
satisfactory analyses cannot be obtained with adjunction. In particular,
using adjunction in this way cannot handle cases in which parts of the
clausal complement are required to be placed within the structure of the
adjoined tree.

The {\DTG} operation of subsertion is designed to overcome this limitation.
Subsertion can be viewed as a generalization of adjunction in which
components of the clausal complement (the subserted structure) which are
not substituted can be interspersed within the structure that is the
site of the subsertion. Following earlier work~\cite{bjr91,vijay92}, {\DTG}
provide a mechanism involving the use of domination links (d-edges) that ensure
that parts of the subserted structure that are not
substituted dominate those parts that are. Furthermore, there is a need
to constrain the way in which the non-substituted components can be
interspersed\footnote{ This was also observed by
Rambow~\shortcite{rambow:1994a}, where an integrity constraint (first
defined for an {\IDLP} version of {\TAG}~\cite{bjr91}) is defined for a
{\MCTAGDL} version called {\VTAG}\@.  However, this was found to be
insufficient for treating both long-distance scrambling and
long-distance topicalization in German.  {\VTAG} retains adjoining (to
handle topicalization) for this reason.}. This is done by either using
appropriate feature constraints at nodes or by means of
subsertion-insertion constraints (see Section~\ref{sec-def}).

We end this section by briefly commenting on the other {\DTG} operation
of sister-adjunction. In {\TAG}, modification is performed with adjunction
of modifier trees that have a highly constrained form. In particular,
the foot nodes of these trees are always daughters of the root and
either the leftmost or rightmost frontier nodes. The effect of adjoining
a tree of this form corresponds (almost) exactly to the addition of a
new (leftmost or rightmost) subtree below the node that was the site of
the adjunction.  For this reason, we have equipped {\DTG} with an
operation (sister-adjunction) that does exactly this and nothing
more. From the definition of {\DTG} in Section~\ref{sec-def} it can be
seen that the essential aspects of Schabes \& Shieber~\shortcite{ss94}
treatment for modification, including multiple modifications of a
phrase, can be captured by using this
operation\footnote{Santorini and
Mahootian~\shortcite{santorini/mahootian:1995} provide additional
evidence against the standard {\TAG} approach to modification from code
switching data, which can be accounted for by using sister-adjunction.}.

After defining {\DTG} in Section~\ref{sec-def}, we discuss, in
Section~\ref{sec-examples}, {\DTG} analyses for the English and Kashmiri
data presented in this section. Section~\ref{sec-properties} briefly
discusses {\DTG} recognition algorithms.

\section{Definition of D-Tree Grammars}
\label{sec-def}

A {\bf d-tree} is a tree with two types of edges: domination edges~({\bf
d-edges}) and immediate domination edges~({\bf i-edges}).  D-edges and
i-edges express domination and immediate domination relations between
nodes. These relations are never rescinded when d-trees are
composed. Thus, nodes separated by an i-edge will remain in a
mother-daughter relationship throughout the derivation, whereas nodes
separated by an d-edge can be equated or have a path of any length
inserted between them during a derivation.  D-edges and i-edges are not
distributed arbitrarily in d-trees.  For each internal node, either all
of its daughters are linked by i-edges or it has a single daughter that
is linked to it by a d-edge.  Each node is labelled with a terminal
symbol, a nonterminal symbol or the empty string.  A d-tree containing
$n$ d-edges can be decomposed into $n+1$ {\bf components} containing
only i-edges.

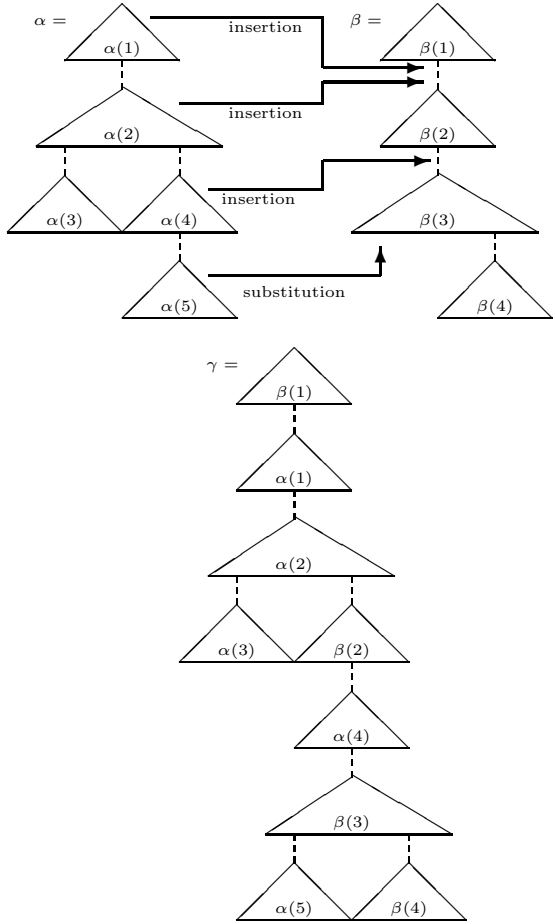
\begin{figure}[bth]
\begin{center}
\setlength{\unitlength}{0.007500in}%
\begingroup\makeatletter\ifx\SetFigFont\undefined
\def\x#1#2#3#4#5#6#7\relax{\def\x{#1#2#3#4#5#6}}%
\expandafter\x\fmtname xxxxxx\relax \def\y{splain}%
\ifx\x\y   
\gdef\SetFigFont#1#2#3{%
  \ifnum #1<17\tiny\else \ifnum #1<20\small\else
  \ifnum #1<24\normalsize\else \ifnum #1<29\large\else
  \ifnum #1<34\Large\else \ifnum #1<41\LARGE\else
     \huge\fi\fi\fi\fi\fi\fi
  \csname #3\endcsname}%
\else
\gdef\SetFigFont#1#2#3{\begingroup
  \count@#1\relax \ifnum 25<\count@\count@25\fi
  \def\x{\endgroup\@setsize\SetFigFont{#2pt}}%
  \expandafter\x
    \csname \romannumeral\the\count@ pt\expandafter\endcsname
    \csname @\romannumeral\the\count@ pt\endcsname
  \csname #3\endcsname}%
\fi
\fi\endgroup
\begin{picture}(400,640)(60,180)
\thinlines
\put( 60,700){\makebox(0.1852,1.2963){\SetFigFont{5}{6}{rm}.}}
\put(380,820){\line(-1,-1){ 40}}
\put(340,780){\line( 1, 0){ 80}}
\put(420,780){\line(-1, 1){ 40}}
\put(160,820){\line(-1,-1){ 40}}
\put(120,780){\line( 1, 0){ 80}}
\put(200,780){\line(-1, 1){ 40}}
\multiput(380,780)(0.00000,-8.00000){3}{\line( 0,-1){  4.000}}
\multiput(160,780)(0.00000,-8.00000){3}{\line( 0,-1){  4.000}}
\put(160,760){\line(-3,-2){ 60}}
\put(100,720){\line( 1, 0){130}}
\put(230,720){\line(-5, 3){ 69.118}}
\put(380,760){\line(-1,-1){ 40}}
\put(340,720){\line( 1, 0){ 80}}
\put(420,720){\line(-1, 1){ 40}}
\multiput(120,720)(0.00000,-8.00000){3}{\line( 0,-1){  4.000}}
\multiput(200,720)(0.00000,-8.00000){3}{\line( 0,-1){  4.000}}
\put(200,700){\line(-1,-1){ 40}}
\put(160,660){\line( 1, 0){ 80}}
\put(240,660){\line(-1, 1){ 40}}
\put(120,700){\line(-1,-1){ 40}}
\put( 80,660){\line( 1, 0){ 80}}
\put(160,660){\line(-1, 1){ 40}}
\multiput(380,720)(0.00000,-8.00000){3}{\line( 0,-1){  4.000}}
\put(200,640){\line(-1,-1){ 40}}
\put(160,600){\line( 1, 0){ 80}}
\put(240,600){\line(-1, 1){ 40}}
\multiput(200,660)(0.00000,-8.00000){3}{\line( 0,-1){  4.000}}
\put(380,700){\line(-3,-2){ 60}}
\put(320,660){\line( 1, 0){130}}
\put(450,660){\line(-5, 3){ 69.118}}
\multiput(420,660)(0.00000,-8.00000){3}{\line( 0,-1){  4.000}}
\put(420,640){\line(-1,-1){ 40}}
\put(380,600){\line( 1, 0){ 80}}
\put(460,600){\line(-1, 1){ 40}}
\thicklines
\put(220,630){\line( 1, 0){120}}
\put(340,630){\vector( 0, 1){ 20}}
\put(220,690){\line( 1, 0){ 80}}
\put(300,690){\line( 0, 1){ 20}}
\put(300,710){\vector( 1, 0){ 75}}
\put(200,750){\line( 1, 0){100}}
\put(300,750){\line( 0, 1){ 15}}
\put(300,765){\vector( 1, 0){ 70}}
\put(180,810){\line( 1, 0){120}}
\put(300,810){\line( 0,-1){ 35}}
\put(300,775){\vector( 1, 0){ 70}}
\thinlines
\put(280,580){\line(-1,-1){ 40}}
\put(240,540){\line( 1, 0){ 80}}
\put(320,540){\line(-1, 1){ 40}}
\put(280,520){\line(-1,-1){ 40}}
\put(240,480){\line( 1, 0){ 80}}
\put(320,480){\line(-1, 1){ 40}}
\multiput(280,540)(0.00000,-8.00000){3}{\line( 0,-1){  4.000}}
\put(280,460){\line(-3,-2){ 60}}
\put(220,420){\line( 1, 0){130}}
\put(350,420){\line(-5, 3){ 69.118}}
\multiput(240,420)(0.00000,-8.00000){3}{\line( 0,-1){  4.000}}
\put(240,400){\line(-1,-1){ 40}}
\put(200,360){\line( 1, 0){ 80}}
\put(280,360){\line(-1, 1){ 40}}
\multiput(320,420)(0.00000,-8.00000){3}{\line( 0,-1){  4.000}}
\put(320,400){\line(-1,-1){ 40}}
\put(280,360){\line( 1, 0){ 80}}
\put(360,360){\line(-1, 1){ 40}}
\multiput(320,360)(0.00000,-8.00000){3}{\line( 0,-1){  4.000}}
\put(320,340){\line(-1,-1){ 40}}
\put(280,300){\line( 1, 0){ 80}}
\put(360,300){\line(-1, 1){ 40}}
\put(320,280){\line(-3,-2){ 60}}
\put(260,240){\line( 1, 0){130}}
\put(390,240){\line(-5, 3){ 69.118}}
\multiput(320,300)(0.00000,-8.00000){3}{\line( 0,-1){  4.000}}
\put(280,220){\line(-1,-1){ 40}}
\put(240,180){\line( 1, 0){ 80}}
\put(320,180){\line(-1, 1){ 40}}
\multiput(280,240)(0.00000,-8.00000){3}{\line( 0,-1){  4.000}}
\multiput(360,240)(0.00000,-8.00000){3}{\line( 0,-1){  4.000}}
\put(360,220){\line(-1,-1){ 40}}
\put(320,180){\line( 1, 0){ 80}}
\put(400,180){\line(-1, 1){ 40}}
\multiput(280,480)(0.00000,-8.00000){3}{\line( 0,-1){  4.000}}
\put(110,805){\makebox(0,0)[b]{\smash{\SetFigFont{6}{7.2}{rm}$\alpha=$}}}
\put(160,785){\makebox(0,0)[b]{\smash{\SetFigFont{6}{7.2}{rm}$\alpha(1)$}}}
\put(380,785){\makebox(0,0)[b]{\smash{\SetFigFont{6}{7.2}{rm}$\beta(1)$}}}
\put(380,725){\makebox(0,0)[b]{\smash{\SetFigFont{6}{7.2}{rm}$\beta(2)$}}}
\put(160,725){\makebox(0,0)[b]{\smash{\SetFigFont{6}{7.2}{rm}$\alpha(2)$}}}
\put(120,665){\makebox(0,0)[b]{\smash{\SetFigFont{6}{7.2}{rm}$\alpha(3)$}}}
\put(200,665){\makebox(0,0)[b]{\smash{\SetFigFont{6}{7.2}{rm}$\alpha(4)$}}}
\put(200,605){\makebox(0,0)[b]{\smash{\SetFigFont{6}{7.2}{rm}$\alpha(5)$}}}
\put(380,665){\makebox(0,0)[b]{\smash{\SetFigFont{6}{7.2}{rm}$\beta(3)$}}}
\put(420,605){\makebox(0,0)[b]{\smash{\SetFigFont{6}{7.2}{rm}$\beta(4)$}}}
\put(255,680){\makebox(0,0)[b]{\smash{\SetFigFont{6}{7.2}{rm}insertion}}}
\put(260,740){\makebox(0,0)[b]{\smash{\SetFigFont{6}{7.2}{rm}insertion}}}
\put(260,800){\makebox(0,0)[b]{\smash{\SetFigFont{6}{7.2}{rm}insertion}}}
\put(280,615){\makebox(0,0)[b]{\smash{\SetFigFont{6}{7.2}{rm}substitution}}}
\put(230,565){\makebox(0,0)[b]{\smash{\SetFigFont{6}{7.2}{rm}$\gamma=$}}}
\put(280,545){\makebox(0,0)[b]{\smash{\SetFigFont{6}{7.2}{rm}$\beta(1)$}}}
\put(280,485){\makebox(0,0)[b]{\smash{\SetFigFont{6}{7.2}{rm}$\alpha(1)$}}}
\put(280,425){\makebox(0,0)[b]{\smash{\SetFigFont{6}{7.2}{rm}$\alpha(2)$}}}
\put(240,365){\makebox(0,0)[b]{\smash{\SetFigFont{6}{7.2}{rm}$\alpha(3)$}}}
\put(320,365){\makebox(0,0)[b]{\smash{\SetFigFont{6}{7.2}{rm}$\beta(2)$}}}
\put(320,305){\makebox(0,0)[b]{\smash{\SetFigFont{6}{7.2}{rm}$\alpha(4)$}}}
\put(280,185){\makebox(0,0)[b]{\smash{\SetFigFont{6}{7.2}{rm}$\alpha(5)$}}}
\put(320,245){\makebox(0,0)[b]{\smash{\SetFigFont{6}{7.2}{rm}$\beta(3)$}}}
\put(360,185){\makebox(0,0)[b]{\smash{\SetFigFont{6}{7.2}{rm}$\beta(4)$}}}
\put(330,805){\makebox(0,0)[b]{\smash{\SetFigFont{6}{7.2}{rm}$\beta=$}}}
\end{picture}
\end{center}
\caption{Subsertion}
\label{fig-subsertion}
\end{figure}

D-trees can be composed using two operations: {\bf subsertion} and {\bf
sister-adjunction}.  When a d-tree $\alpha$ is subserted into another
d-tree $\beta$, a component of $\alpha$ is substituted at a frontier
nonterminal node (a {\bf substitution node}) of $\beta$ and all
components of $\alpha$ that are above the substituted component are
inserted into d-edges above the substituted node or placed above the
root node. For example, consider the d-trees $\alpha$ and $\beta$ shown
in Figure~\ref{fig-subsertion}. Note that components are shown as
triangles.  In the composed d-tree $\gamma$ the component $\alpha(5)$ is
substituted at a substitution node in $\beta$.  The components,
$\alpha(1)$, $\alpha(2)$, and $\alpha(4)$ of $\alpha$
above $\alpha(5)$ drift up the path in $\beta$ which runs from the
substitution node.  These components are then {\bf inserted} into
d-edges in $\beta$ or above the root of $\beta$.  In general, when a
component $\alpha(i)$ of some d-tree $\alpha$ is inserted into a d-edge
between nodes $\eta_1$ and $\eta_2$ two new d-edges are created, the
first of which relates $\eta_1$ and the root node of $\alpha(i)$, and
the second of which relates the frontier node of $\alpha(i)$ that
dominates the substituted component to $\eta_2$.  It is possible for
components above the substituted node to drift arbitrarily far up the
d-tree and distribute themselves within domination edges, or above the
root, in any way that is compatible with the domination relationships
present in the substituted d-tree. {\DTG} provide a mechanism
called {\bf subsertion-insertion constraints} to control what can appear
within d-edges (see below).

The second composition operation involving d-trees is called
sister-adjunction. When a d-tree $\alpha$ is sister-adjoined at a node
$\eta$ in a d-tree $\beta$ the composed d-tree $\gamma$ results from the
addition to $\beta$ of $\alpha$ as a new leftmost or rightmost
sub-d-tree below $\eta$. Note that sister-adjunction involves the
addition of exactly one new immediate domination edge and that several
sister-adjunctions can occur at the same node. {\bf Sister-adjoining
constraints} specify where d-trees can be sister-adjoined and whether
they will be right- or left-sister-adjoined (see below).

A {\DTG} is a four tuple $G=(V_N,V_T,S,D)$ where $V_N$ and $V_T$ are
the usual nonterminal and terminal alphabets, $S\in V_N$ is a
distinguished nonterminal and $D$ is a finite set of {\bf elementary}
d-trees. A {\DTG} is said to be {\bf lexicalized} if each d-tree in the
grammar has at least one terminal node.  The elementary d-trees of a
grammar $G$ have two additional annotations: subsertion-insertion
constraints and sister-adjoining constraints. These will be described
below, but first we define simultaneously {\DTG} derivations and
subsertion-adjoining trees~({\SAtree}s), which are partial derivation
structures that can be interpreted as representing dependency
information, the importance of which was stressed in the
introduction\protect\footnote{Due to space limitations, in the following
definitions we are forced to be somewhat imprecise when we identify a
node in a derived d-tree with the node in the elementary d-trees
(elementary nodes) from which it was derived. This is often done in
{\TAG} literature, and hopefully it will be clear what is intended.}.

Consider a {\DTG} $G=(V_N,V_T,S,D)$. In defining {\SAtree}s, we assume some
naming convention for the elementary d-trees in $D$ and some consistent
ordering on the components and nodes of elementary d-trees in $D$.  For
each $i$, we define the set of d-trees $T_i(G)$ whose derivations are
captured by {\SAtree}s of height $i$ or less.  Let $T_0(G)$ be the set $D$
of elementary d-trees of $G$.  Mark all of the components of each d-tree
in $T_0(G)$ as being {\bf substitutable}\protect\footnote{We will
discuss the notion of substitutability further in the next section.  It
is used to ensure the {\SAtree} is a tree. That is, an elementary
structure cannot be subserted into more than one structure since this
would be counter to our motivations for using subsertion for
complementation.}.  Only components marked as substitutable can be
substituted in a subsertion operation.  The {\SAtree} for $\alpha\in
T_0(G)$ consists of a single node labelled by the elementary d-tree name
for $\alpha$.

For $i>0$ let $T_i(G)$ be the union of the set $T_{i-1}(G)$ with the set
of all d-trees $\gamma$ that can be produced as follows. Let $\alpha\in
D$ and let $\gamma$ be the result of subserting or sister-adjoining the
d-trees $\gamma_1,\ldots,\gamma_k$ into $\alpha$ where
$\gamma_1,\ldots,\gamma_k$ are all in $T_{i-1}(G)$, with the subsertions
taking place at different substitution nodes in $\alpha$ as the
footnote.  Only substitutable components of $\gamma_1,\ldots,\gamma_k$
can be substituted in these subsertions. Only the new components of
$\gamma$ that came from $\alpha$ are marked as substitutable in
$\gamma$. Let $\tau_1,\ldots,\tau_k$ be the {\SAtree}s for
$\gamma_1,\ldots,\gamma_k$, respectively.  The {\SAtree} $\tau$ for
$\gamma$ has root labelled by the name for $\alpha$ and $k$ subtrees
$\tau_1,\ldots,\tau_k$. The edge from the root of $\tau$ to the root of
the subtree $\tau_i$ is labelled by $l_i$ ($1\le i\le k$) defined as
follows.  Suppose that $\gamma_i$ was subserted into $\alpha$ and the
root of $\tau_i$ is labelled by the name of some $\alpha'\in D$. Only
components of $\alpha'$ will have been marked as substitutable in
$\gamma_i$. Thus, in this subsertion some component $\alpha'(j)$ will
have been substituted at a node in $\alpha$ with address $n$. In this
case, the label $l_i$ is the pair $(j,n)$.  Alternatively, $\gamma_i$
will have been $d$-sister-adjoined at some node with address $n$ in
$\alpha$, in which case $l_i$ will be the pair $(d,n)$ where
$d\in\set{{\sf left},{\sf right}}$.

The {\bf tree set} $T(G)$ generated by $G$ is defined as the set of
trees $\gamma$ such that: $\gamma'\in T_i(G)$ for some $i\ge 0$;
$\gamma'$ is rooted with the nonterminal $S$; the frontier of $\gamma'$
is a string in $V_T^*$; and $\gamma$ results from the removal of all
d-edges from $\gamma'$.  A d-edge is removed by merging the nodes at
either end of the edge as long as they are labelled by the same symbol.
The {\bf string language} $L(G)$ associated with $G$ is the set of
terminal strings appearing on the frontier of trees in $T(G)$.

We have given a reasonably precise definition of {\SAtree}s since
they play such an important role in the motivation for this work.  We
now describe informally a structure that can be used to encode a {\DTG}
derivation. A derivation graph for $\gamma\in T(G)$ results from the
addition of insertion edges to a {\SAtree} $\tau$ for $\gamma$.
The location in $\gamma$ of an inserted elementary component $\alpha(i)$
can be unambiguously determined by identifying the source of the node
(say the node with address $n$ in the elementary d-tree $\alpha'$) with
which the root of this occurrence of $\alpha(i)$ is merged with when
d-edges are removed. The insertion edge will relate the two (not
necessarily distinct) nodes corresponding to appropriate occurrences of
$\alpha$ and $\alpha'$ and will be labelled by the pair $(i,n)$.

Each d-edge in elementary d-trees has an associated subsertion-insertion
constraint~({\SIC}). A {\SIC} is a finite set of elementary node
addresses~({\ENA}s). An {\ENA} $\eta$ specifies some elementary d-tree
$\alpha\in D$, a component of $\alpha$ and the address of a node within
that component of $\alpha$. If a {\ENA} $\eta$ is in the {\SIC} associated
with a d-edge between $\eta_1$ and $\eta_2$ in an elementary d-tree
$\alpha$ then $\eta$ cannot appear properly
within the path that appears from
$\eta_1$ to $\eta_2$ in the derived tree $\gamma\in T(G)$.

Each node of elementary d-trees has an associated sister-adjunction
constraint~({\SAC}). A {\SAC} is a finite set of pairs, each pair
identifying a direction (left or right) and an elementary d-tree. A
{\SAC} gives a complete specification of what can be sister-adjoined at
a node.  If a node $\eta$ is associated with a {\SAC} containing a pair
$(d,\alpha)$ then the d-tree $\alpha$ can be $d$-sister-adjoined at
$\eta$. By definition of sister-adjunction, all substitution nodes and
all nodes at the top of d-edges can be assumed to have {\SAC}s that are
the empty-set. This prevents sister-adjunction at these nodes.

In this section we have defined ``raw'' {\DTG}. In a more refined
version of the formalism we would associate (a single) finite-valued
feature structure with each node\protect\footnote{Trees used in
Section~\ref{sec-examples} make use of such feature structures.}.  It is
a matter of further research to determine to what extent {\SIC}s and
{\SAC}s can be stated globally for a grammar, rather than being attached
to d-edges/nodes\protect\footnote{In this context, it might be beneficial
to consider
the expression of a feature-based lexicalist theory such as
{\HPSG} in DTG, similar to the compilation of {\HPSG} to
{\TAG}~\cite{kknv95}.}.  See the next section for a brief discussion of
linguistic principles from which a grammar's {\SIC}s could be derived.

\section{Linguistic Examples}

\label{sec-examples}

In this section, we show how an account for the data introduced in
Section~\ref{sec-intro} can be given with {\DTG}\@.

\subsection{Getting Dependencies Right: English}

\begin{figure}[htb]
\begin{center}
\setlength{\unitlength}{0.008125in}%
\begingroup\makeatletter\ifx\SetFigFont\undefined
\def\x#1#2#3#4#5#6#7\relax{\def\x{#1#2#3#4#5#6}}%
\expandafter\x\fmtname xxxxxx\relax \def\y{splain}%
\ifx\x\y   
\gdef\SetFigFont#1#2#3{%
  \ifnum #1<17\tiny\else \ifnum #1<20\small\else
  \ifnum #1<24\normalsize\else \ifnum #1<29\large\else
  \ifnum #1<34\Large\else \ifnum #1<41\LARGE\else
     \huge\fi\fi\fi\fi\fi\fi
  \csname #3\endcsname}%
\else
\gdef\SetFigFont#1#2#3{\begingroup
  \count@#1\relax \ifnum 25<\count@\count@25\fi
  \def\x{\endgroup\@setsize\SetFigFont{#2pt}}%
  \expandafter\x
    \csname \romannumeral\the\count@ pt\expandafter\endcsname
    \csname @\romannumeral\the\count@ pt\endcsname
  \csname #3\endcsname}%
\fi
\fi\endgroup
\begin{picture}(129,213)(113,565)
\thinlines
\multiput(160,760)(0.00000,-8.00000){3}{\line( 0,-1){  4.000}}
\put(120,700){\line( 2, 1){ 40}}
\put(160,720){\line( 2,-1){ 40}}
\multiput(200,680)(0.00000,-8.00000){3}{\line( 0,-1){  4.000}}
\put(160,620){\line( 2, 1){ 40}}
\put(200,640){\line( 2,-1){ 40}}
\put(160,600){\line( 0,-1){ 20}}
\put(160,765){\makebox(0,0)[b]{\smash{\SetFigFont{7}{8.4}{rm}S'}}}
\put(210,685){\makebox(0,0)[lb]{\smash{\SetFigFont{8}{9.6}{rm}[fin: +]}}}
\put(160,725){\makebox(0,0)[b]{\smash{\SetFigFont{7}{8.4}{rm}S}}}
\put(120,685){\makebox(0,0)[b]{\smash{\SetFigFont{7}{8.4}{rm}NP}}}
\put(200,685){\makebox(0,0)[b]{\smash{\SetFigFont{7}{8.4}{rm}VP}}}
\put(210,645){\makebox(0,0)[lb]{\smash{\SetFigFont{8}{9.6}{rm}[fin: +]}}}
\put(200,645){\makebox(0,0)[b]{\smash{\SetFigFont{7}{8.4}{rm}VP}}}
\put(160,605){\makebox(0,0)[b]{\smash{\SetFigFont{7}{8.4}{rm}V}}}
\put(160,565){\makebox(0,0)[b]{\smash{\SetFigFont{7}{8.4}{rm}claims}}}
\put(240,605){\makebox(0,0)[b]{\smash{\SetFigFont{7}{8.4}{rm}S}}}
\end{picture}
\qquad
\setlength{\unitlength}{0.008125in}%
\begingroup\makeatletter\ifx\SetFigFont\undefined
\def\x#1#2#3#4#5#6#7\relax{\def\x{#1#2#3#4#5#6}}%
\expandafter\x\fmtname xxxxxx\relax \def\y{splain}%
\ifx\x\y   
\gdef\SetFigFont#1#2#3{%
  \ifnum #1<17\tiny\else \ifnum #1<20\small\else
  \ifnum #1<24\normalsize\else \ifnum #1<29\large\else
  \ifnum #1<34\Large\else \ifnum #1<41\LARGE\else
     \huge\fi\fi\fi\fi\fi\fi
  \csname #3\endcsname}%
\else
\gdef\SetFigFont#1#2#3{\begingroup
  \count@#1\relax \ifnum 25<\count@\count@25\fi
  \def\x{\endgroup\@setsize\SetFigFont{#2pt}}%
  \expandafter\x
    \csname \romannumeral\the\count@ pt\expandafter\endcsname
    \csname @\romannumeral\the\count@ pt\endcsname
  \csname #3\endcsname}%
\fi
\fi\endgroup
\begin{picture}(102,133)(88,645)
\thinlines
\multiput(140,760)(0.00000,-8.00000){3}{\line( 0,-1){  4.000}}
\put(100,700){\line( 2, 1){ 40}}
\put(140,720){\line( 2,-1){ 40}}
\put(100,680){\line( 0,-1){ 20}}
\put(140,765){\makebox(0,0)[b]{\smash{\SetFigFont{7}{8.4}{rm}S}}}
\put(190,685){\makebox(0,0)[lb]{\smash{\SetFigFont{8}{9.6}{rm}[fin: -]}}}
\put(150,725){\makebox(0,0)[lb]{\smash{\SetFigFont{8}{9.6}{rm}[fin: +]}}}
\put(140,725){\makebox(0,0)[b]{\smash{\SetFigFont{7}{8.4}{rm}VP}}}
\put(180,685){\makebox(0,0)[b]{\smash{\SetFigFont{7}{8.4}{rm}VP}}}
\put(100,645){\makebox(0,0)[b]{\smash{\SetFigFont{7}{8.4}{rm}seems}}}
\put(100,685){\makebox(0,0)[b]{\smash{\SetFigFont{7}{8.4}{rm}V}}}
\end{picture}
\setlength{\unitlength}{0.008125in}%
\begingroup\makeatletter\ifx\SetFigFont\undefined
\def\x#1#2#3#4#5#6#7\relax{\def\x{#1#2#3#4#5#6}}%
\expandafter\x\fmtname xxxxxx\relax \def\y{splain}%
\ifx\x\y   
\gdef\SetFigFont#1#2#3{%
  \ifnum #1<17\tiny\else \ifnum #1<20\small\else
  \ifnum #1<24\normalsize\else \ifnum #1<29\large\else
  \ifnum #1<34\Large\else \ifnum #1<41\LARGE\else
     \huge\fi\fi\fi\fi\fi\fi
  \csname #3\endcsname}%
\else
\gdef\SetFigFont#1#2#3{\begingroup
  \count@#1\relax \ifnum 25<\count@\count@25\fi
  \def\x{\endgroup\@setsize\SetFigFont{#2pt}}%
  \expandafter\x
    \csname \romannumeral\the\count@ pt\expandafter\endcsname
    \csname @\romannumeral\the\count@ pt\endcsname
  \csname #3\endcsname}%
\fi
\fi\endgroup
\begin{picture}(188,253)(99,485)
\thinlines
\put(120,700){\line( 2, 1){ 40}}
\put(160,720){\line( 2,-1){ 40}}
\multiput(200,680)(0.00000,-8.00000){3}{\line( 0,-1){  4.000}}
\put(160,620){\line( 2, 1){ 40}}
\put(200,640){\line( 2,-1){ 40}}
\multiput(240,600)(0.00000,-8.00000){3}{\line( 0,-1){  4.000}}
\put(200,540){\line( 2, 1){ 40}}
\put(240,560){\line( 2,-1){ 40}}
\put(200,520){\line( 0,-1){ 20}}
\put(280,520){\line( 0,-1){ 20}}
\put(160,725){\makebox(0,0)[b]{\smash{\SetFigFont{7}{8.4}{rm}S'}}}
\put(120,685){\makebox(0,0)[b]{\smash{\SetFigFont{7}{8.4}{rm}NP}}}
\put(200,685){\makebox(0,0)[b]{\smash{\SetFigFont{7}{8.4}{rm}S}}}
\put(120,665){\makebox(0,0)[b]{\smash{\SetFigFont{7}{8.4}{rm}(hotdogs)}}}
\put(200,645){\makebox(0,0)[b]{\smash{\SetFigFont{7}{8.4}{rm}S}}}
\put(160,605){\makebox(0,0)[b]{\smash{\SetFigFont{7}{8.4}{rm}NP}}}
\put(240,605){\makebox(0,0)[b]{\smash{\SetFigFont{7}{8.4}{rm}VP}}}
\put(160,585){\makebox(0,0)[b]{\smash{\SetFigFont{7}{8.4}{rm}(Mary)}}}
\put(240,565){\makebox(0,0)[b]{\smash{\SetFigFont{7}{8.4}{rm}VP}}}
\put(200,485){\makebox(0,0)[b]{\smash{\SetFigFont{7}{8.4}{rm}to adore}}}
\put(200,525){\makebox(0,0)[b]{\smash{\SetFigFont{7}{8.4}{rm}V}}}
\put(280,525){\makebox(0,0)[b]{\smash{\SetFigFont{7}{8.4}{rm}NP}}}
\put(280,485){\makebox(0,0)[b]{\smash{\SetFigFont{7}{8.4}{rm}e}}}
\put(250,565){\makebox(0,0)[lb]{\smash{\SetFigFont{8}{9.6}{rm}[fin: -]}}}
\put(250,605){\makebox(0,0)[lb]{\smash{\SetFigFont{8}{9.6}{rm}[fin: +]}}}
\end{picture}
\end{center}
\caption{D-trees for (\protect\ref{exa-english})}
\label{fig-english}
\end{figure}
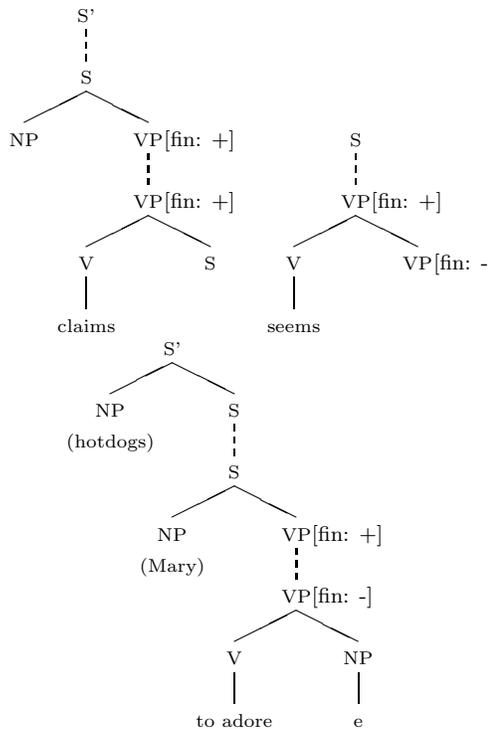

In Figure~\ref{fig-english}, we give a {\DTG} that generates sentence
(\ref{exa-english}).  Every d-tree is a projection from a lexical
anchor.  The label of the maximal projection is, we assume, determined
by the morphology of the anchor.  For example, if the anchor is a finite
verb, it will project to S, indicating that an overt syntactic
(``surface'') subject is required for agreement with it (and perhaps
case-assignment).  Furthermore, a finite verb may optionally also
project to S$'$ (as in the d-tree shown for {\em claims}), indicating
that a {\em wh}-moved or topicalized element is required.  The finite
verb {\em seems} also projects to S, even though it does not itself
provide a functional subject.  In the case of the {\em to adore} tree,
the situation is the inverse: the functional subject requires a finite
verb to agree with, which is signaled by the fact that its component's
root and frontier nodes are labelled S and VP, respectively, but the
verb itself is not finite and therefore only projects to VP[-{\featf
fin}].  Therefore, the subject will have to raise out of its clause for
agreement and case assignment.  The direct object of {\em to adore} has
{\em wh-}moved out of the projection of the verb (we include a trace for
the sake of clarity).

\begin{figure}[htb]
\begin{center}
\setlength{\unitlength}{0.008125in}%
\begingroup\makeatletter\ifx\SetFigFont\undefined
\def\x#1#2#3#4#5#6#7\relax{\def\x{#1#2#3#4#5#6}}%
\expandafter\x\fmtname xxxxxx\relax \def\y{splain}%
\ifx\x\y   
\gdef\SetFigFont#1#2#3{%
  \ifnum #1<17\tiny\else \ifnum #1<20\small\else
  \ifnum #1<24\normalsize\else \ifnum #1<29\large\else
  \ifnum #1<34\Large\else \ifnum #1<41\LARGE\else
     \huge\fi\fi\fi\fi\fi\fi
  \csname #3\endcsname}%
\else
\gdef\SetFigFont#1#2#3{\begingroup
  \count@#1\relax \ifnum 25<\count@\count@25\fi
  \def\x{\endgroup\@setsize\SetFigFont{#2pt}}%
  \expandafter\x
    \csname \romannumeral\the\count@ pt\expandafter\endcsname
    \csname @\romannumeral\the\count@ pt\endcsname
  \csname #3\endcsname}%
\fi
\fi\endgroup
\begin{picture}(299,293)(208,525)
\thinlines
\put(260,680){\line( 0,-1){ 20}}
\put(260,640){\line( 0,-1){ 20}}
\put(260,685){\makebox(0,0)[b]{\smash{\SetFigFont{7}{8.4}{rm}AdjP}}}
\put(260,645){\makebox(0,0)[b]{\smash{\SetFigFont{7}{8.4}{rm}Adj}}}
\put(260,605){\makebox(0,0)[b]{\smash{\SetFigFont{7}{8.4}{rm}spicy}}}
\put(220,680){\line( 0,-1){ 20}}
\put(220,640){\line( 0,-1){ 20}}
\put(220,685){\makebox(0,0)[b]{\smash{\SetFigFont{7}{8.4}{rm}AdjP}}}
\put(220,645){\makebox(0,0)[b]{\smash{\SetFigFont{7}{8.4}{rm}Adj}}}
\put(220,605){\makebox(0,0)[b]{\smash{\SetFigFont{7}{8.4}{rm}small}}}
\put(300,680){\line( 0,-1){ 20}}
\put(300,685){\makebox(0,0)[b]{\smash{\SetFigFont{7}{8.4}{rm}N}}}
\put(300,645){\makebox(0,0)[b]{\smash{\SetFigFont{7}{8.4}{rm}hotdogs}}}
\put(460,560){\line( 0,-1){ 20}}
\put(460,525){\makebox(0,0)[b]{\smash{\SetFigFont{7}{8.4}{rm}to adore}}}
\put(460,565){\makebox(0,0)[b]{\smash{\SetFigFont{7}{8.4}{rm}V}}}
\put(500,560){\line( 0,-1){ 20}}
\put(500,565){\makebox(0,0)[b]{\smash{\SetFigFont{7}{8.4}{rm}NP}}}
\put(500,525){\makebox(0,0)[b]{\smash{\SetFigFont{7}{8.4}{rm}e}}}
\put(260,780){\line( 2, 1){ 40}}
\put(300,800){\line( 3,-1){ 60}}
\put(320,740){\line( 2, 1){ 40}}
\put(360,760){\line( 2,-1){ 40}}
\put(360,700){\line( 2, 1){ 40}}
\put(400,720){\line( 2,-1){ 40}}
\put(360,680){\line( 0,-1){ 20}}
\put(400,660){\line( 2, 1){ 40}}
\put(440,680){\line( 2,-1){ 40}}
\put(320,720){\line( 0,-1){ 20}}
\put(400,640){\line( 0,-1){ 20}}
\put(260,760){\line( 0,-1){ 20}}
\put(260,720){\line( 0,-1){ 20}}
\put(260,720){\line(-2,-1){ 40}}
\put(260,720){\line( 2,-1){ 40}}
\put(440,620){\line( 2, 1){ 40}}
\put(480,640){\line( 0,-1){ 20}}
\put(460,580){\line( 1, 1){ 20}}
\put(480,600){\line( 1,-1){ 20}}
\put(300,805){\makebox(0,0)[b]{\smash{\SetFigFont{7}{8.4}{rm}S'}}}
\put(360,765){\makebox(0,0)[b]{\smash{\SetFigFont{7}{8.4}{rm}S}}}
\put(320,725){\makebox(0,0)[b]{\smash{\SetFigFont{7}{8.4}{rm}NP}}}
\put(400,725){\makebox(0,0)[b]{\smash{\SetFigFont{7}{8.4}{rm}VP}}}
\put(360,685){\makebox(0,0)[b]{\smash{\SetFigFont{7}{8.4}{rm}V}}}
\put(360,645){\makebox(0,0)[b]{\smash{\SetFigFont{7}{8.4}{rm}claims}}}
\put(440,685){\makebox(0,0)[b]{\smash{\SetFigFont{7}{8.4}{rm}S}}}
\put(400,645){\makebox(0,0)[b]{\smash{\SetFigFont{7}{8.4}{rm}NP}}}
\put(320,685){\makebox(0,0)[b]{\smash{\SetFigFont{7}{8.4}{rm}he}}}
\put(400,605){\makebox(0,0)[b]{\smash{\SetFigFont{7}{8.4}{rm}Mary}}}
\put(480,645){\makebox(0,0)[b]{\smash{\SetFigFont{7}{8.4}{rm}VP}}}
\put(440,605){\makebox(0,0)[b]{\smash{\SetFigFont{7}{8.4}{rm}seems}}}
\put(260,725){\makebox(0,0)[b]{\smash{\SetFigFont{7}{8.4}{rm}N'}}}
\put(260,765){\makebox(0,0)[b]{\smash{\SetFigFont{7}{8.4}{rm}NP}}}
\put(480,605){\makebox(0,0)[b]{\smash{\SetFigFont{7}{8.4}{rm}VP}}}
\end{picture}
\end{center}
\caption{Derived tree for (\protect\ref{exa-english})}
\label{fig-eng-derived}
\end{figure}
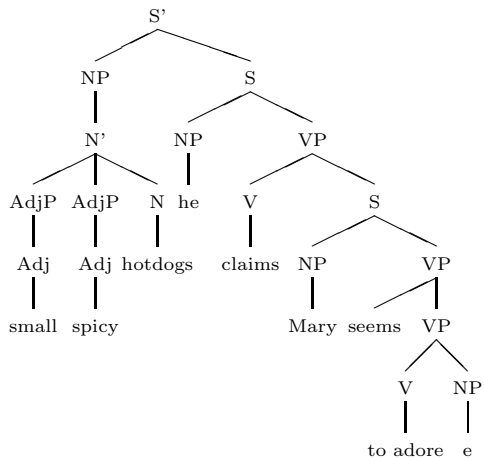

We add {\SIC}s to ensure that the projections are respected by components
of other d-trees that may be inserted during a derivation.  A {\SIC} is
associated with the d-edge between VP and S node in the {\em seems}
d-tree to ensure that no node labelled S$'$ can be inserted within it --
i.e., it can not be filled by with a {\em wh}-moved element.  In
contrast, since both the subject and the object of {\em to adore} have
been moved out of the projection of the verb, the path to these
arguments do not carry any {\SIC} at all\footnote{We enforce island
effects for {\em wh}-movement by using a [$\pm${\featf extract}] feature
on substitution nodes.  This corresponds roughly to the analysis in
{\TAG}, where islandhood is (to a large extent) enforced by designating a
particular node as the foot node \cite{kj86}.}.

We now discuss a possible derivation.  We start out with the most deeply
embedded clause, the {\em adores} clause.  Before subserting its nominal
arguments, we sister-adjoin the two adjectival trees to the tree for
{\em hotdogs}.  This is handled by a {\SAC} associated with the N$'$
node that allows all trees rooted in AdjP to be left sister-adjoined.
We then subsert this structure and the subject into the {\em to adore}
d-tree.  We subsert the resulting structure into the {\em seems} clause
by substituting its maximal projection node, labelled VP[{\featf fin}:
-], at the VP[{\featf fin}: -] frontier node of {\em seems}, and by
inserting the subject into the d-edge of the {\em seems} tree.  Now,
only the S node of the {\em seems} tree (which is its maximal
projection) is substitutable.  Finally, we subsert this derived
structure into the {\em claims} d-tree by substituting the S node of
{\em seems} at the S complement node of {\em claims}, and by inserting
the object of {\em adores} (which has not yet been used in the
derivation) in the d-edge of the {\em claims} d-tree above its S node.
The derived tree is shown in Figure~\ref{fig-eng-derived}.  The
{\SAtree} for this derivation corresponds to the dependency tree given
previously in Figure~\ref{fig-dep}.

\begin{figure}[htb]
\begin{center}
\setlength{\unitlength}{0.008125in}%
\begingroup\makeatletter\ifx\SetFigFont\undefined
\def\x#1#2#3#4#5#6#7\relax{\def\x{#1#2#3#4#5#6}}%
\expandafter\x\fmtname xxxxxx\relax \def\y{splain}%
\ifx\x\y   
\gdef\SetFigFont#1#2#3{%
  \ifnum #1<17\tiny\else \ifnum #1<20\small\else
  \ifnum #1<24\normalsize\else \ifnum #1<29\large\else
  \ifnum #1<34\Large\else \ifnum #1<41\LARGE\else
     \huge\fi\fi\fi\fi\fi\fi
  \csname #3\endcsname}%
\else
\gdef\SetFigFont#1#2#3{\begingroup
  \count@#1\relax \ifnum 25<\count@\count@25\fi
  \def\x{\endgroup\@setsize\SetFigFont{#2pt}}%
  \expandafter\x
    \csname \romannumeral\the\count@ pt\expandafter\endcsname
    \csname @\romannumeral\the\count@ pt\endcsname
  \csname #3\endcsname}%
\fi
\fi\endgroup
\begin{picture}(283,393)(139,405)
\thinlines
\put(160,660){\line( 2, 1){ 40}}
\put(200,680){\line( 2,-1){ 40}}
\multiput(240,640)(0.00000,-8.00000){3}{\line( 0,-1){  4.000}}
\multiput(240,600)(0.00000,-8.00000){3}{\line( 0,-1){  4.000}}
\put(200,540){\line( 2, 1){ 40}}
\put(240,560){\line( 2,-1){ 40}}
\put(241,503){\line( 2, 1){ 40}}
\put(281,523){\line( 2,-1){ 40}}
\put(280,460){\line( 2, 1){ 40}}
\put(320,480){\line( 2,-1){ 40}}
\put(280,440){\line( 0,-1){ 20}}
\put(235,480){\line( 0,-1){ 20}}
\put(360,440){\line( 0,-1){ 20}}
\put(200,685){\makebox(0,0)[b]{\smash{\SetFigFont{7}{8.4}{rm}S'}}}
\put(160,645){\makebox(0,0)[b]{\smash{\SetFigFont{7}{8.4}{rm}NP}}}
\put(240,645){\makebox(0,0)[b]{\smash{\SetFigFont{7}{8.4}{rm}S}}}
\put(160,625){\makebox(0,0)[b]{\smash{\SetFigFont{7}{8.4}{rm}(hotdogs)}}}
\put(240,605){\makebox(0,0)[b]{\smash{\SetFigFont{7}{8.4}{rm}S}}}
\put(235,485){\makebox(0,0)[b]{\smash{\SetFigFont{7}{8.4}{rm}V}}}
\put(280,405){\makebox(0,0)[b]{\smash{\SetFigFont{7}{8.4}{rm}to adore}}}
\put(280,445){\makebox(0,0)[b]{\smash{\SetFigFont{7}{8.4}{rm}V}}}
\put(360,445){\makebox(0,0)[b]{\smash{\SetFigFont{7}{8.4}{rm}NP}}}
\put(360,405){\makebox(0,0)[b]{\smash{\SetFigFont{7}{8.4}{rm}e}}}
\put(320,485){\makebox(0,0)[b]{\smash{\SetFigFont{7}{8.4}{rm}VP}}}
\put(240,565){\makebox(0,0)[b]{\smash{\SetFigFont{7}{8.4}{rm}S}}}
\put(200,525){\makebox(0,0)[b]{\smash{\SetFigFont{7}{8.4}{rm}NP}}}
\put(280,525){\makebox(0,0)[b]{\smash{\SetFigFont{7}{8.4}{rm}VP}}}
\put(330,485){\makebox(0,0)[lb]{\smash{\SetFigFont{7}{8.4}{rm}[fin: -]}}}
\put(290,525){\makebox(0,0)[lb]{\smash{\SetFigFont{7}{8.4}{rm}[fin: +]}}}
\put(200,505){\makebox(0,0)[b]{\smash{\SetFigFont{7}{8.4}{rm}(Mary)}}}
\put(235,445){\makebox(0,0)[b]{\smash{\SetFigFont{7}{8.4}{rm}seems}}}
\multiput(340,780)(0.00000,-8.00000){3}{\line( 0,-1){  4.000}}
\put(300,720){\line( 2, 1){ 40}}
\put(340,740){\line( 2,-1){ 40}}
\multiput(380,700)(0.00000,-8.00000){3}{\line( 0,-1){  4.000}}
\put(340,640){\line( 2, 1){ 40}}
\put(380,660){\line( 2,-1){ 40}}
\put(340,620){\line( 0,-1){ 20}}
\thicklines
\put(250,570){\line( 1, 0){170}}
\put(420,570){\vector( 0, 1){ 40}}
\put(250,610){\line( 1, 0){ 30}}
\put(280,610){\line( 0, 1){155}}
\put(280,765){\vector( 1, 0){ 50}}
\put(210,690){\line( 1, 0){ 50}}
\put(260,690){\line( 0, 1){ 85}}
\put(260,775){\vector( 1, 0){ 70}}
\put(340,785){\makebox(0,0)[b]{\smash{\SetFigFont{7}{8.4}{rm}S'}}}
\put(340,745){\makebox(0,0)[b]{\smash{\SetFigFont{7}{8.4}{rm}S}}}
\put(300,705){\makebox(0,0)[b]{\smash{\SetFigFont{7}{8.4}{rm}NP}}}
\put(380,705){\makebox(0,0)[b]{\smash{\SetFigFont{7}{8.4}{rm}VP}}}
\put(390,705){\makebox(0,0)[lb]{\smash{\SetFigFont{8}{9.6}{rm}[fin: +]}}}
\put(390,665){\makebox(0,0)[lb]{\smash{\SetFigFont{8}{9.6}{rm}[fin: +]}}}
\put(380,665){\makebox(0,0)[b]{\smash{\SetFigFont{7}{8.4}{rm}VP}}}
\put(340,625){\makebox(0,0)[b]{\smash{\SetFigFont{7}{8.4}{rm}V}}}
\put(340,585){\makebox(0,0)[b]{\smash{\SetFigFont{7}{8.4}{rm}claims}}}
\put(420,625){\makebox(0,0)[b]{\smash{\SetFigFont{7}{8.4}{rm}S}}}
\put(270,785){\makebox(0,0)[lb]{\smash{\SetFigFont{7}{8.4}{rm}Insertions}}}
\put(385,555){\makebox(0,0)[rb]{\smash{\SetFigFont{7}{8.4}{rm}Substitution}}}
\end{picture}
\end{center}
\caption{An ill-formed derivation}
\label{fig-badderiv}
\end{figure}
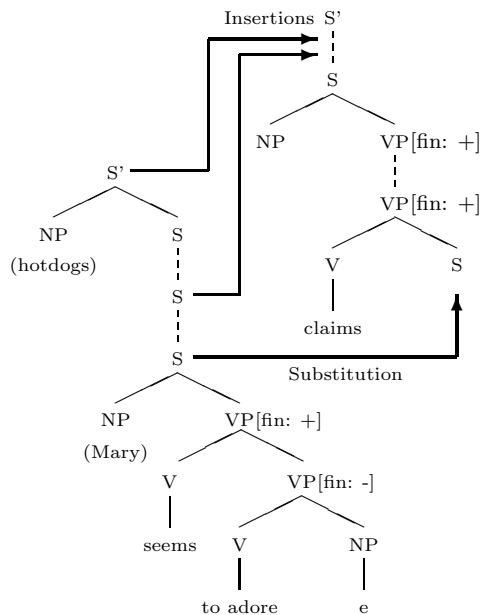

Note that this is the only possible derivation involving these three
d-trees, modulo order of operations.  To see this, consider the
following putative alternate derivation.  We first subsert the {\em to
adore} d-tree into the {\em seems} tree as above, by substituting the
anchor component at the substitution node of {\em seems}.  We insert the
subject component of {\em to adore} above the anchor component of {\em
seems}.  We then subsert this derived structure into the {\em claims}
tree by substituting the root of the subject component of {\em to adore}
at the S node of {\em claims} and by inserting the S node of the {\em
seems} d-tree as well as the object component of the {\em to adore}
d-tree in the S$'$/S d-edge of the {\em claims} d-tree.  This last
operation is shown in Figure~\ref{fig-badderiv}.  The resulting phrase
structure tree would be the same as in the previously discussed
derivation, but the derivation structure is linguistically meaningless,
since {\em to adore} would have been subserted into both {\em seems} and
{\em claims}.  However, this derivation is ruled out by the restriction
that only substitutable components can be substituted: the subject
component of the {\em adore} d-tree is not substitutable after
subsertion into the {\em seems} d-tree, and therefore it cannot be
substituted into the {\em claims} d-tree.

In the above discussion, substitutability played a central role in
ruling out the derivation. We observe in passing that the {\SIC} associated
to the d-edge in the {\em seems} d-tree also rules out this
derivation. The derivation requires that the S node of {\em seems} be
inserted into the S$'$/S d-edge of {\em claims}.  However, we would have
to stretch the edge over two components which are both ruled out by the
{\SIC}, since they violate the projection from {\em seems} to its S node.
Thus, the derivation is excluded by the independently motivated {\SIC}s,
which enforce the notion of projection.  This raises the possibility
that, in grammars that express certain linguistic principles,
substitutability is not needed for ruling out derivations of this
nature.  We intend to examine this issue in future work.

\subsection{Getting Word Order Right: Kashmiri}

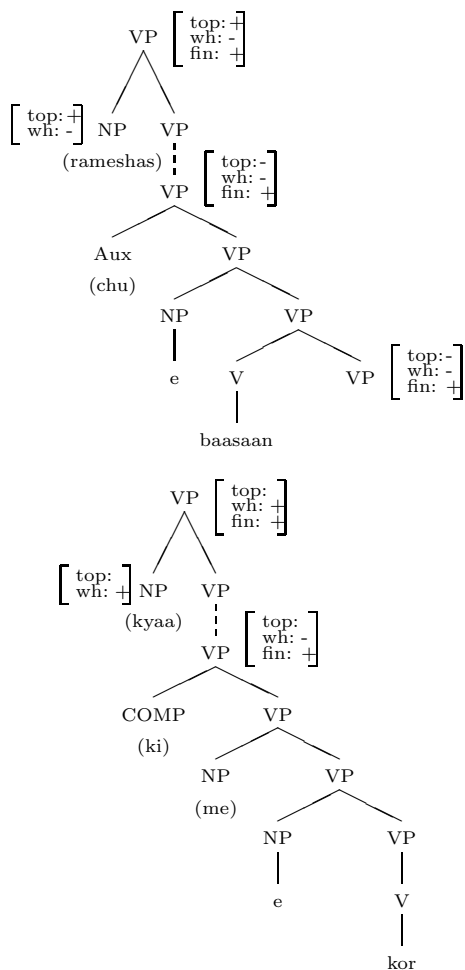
\begin{figure}[htb]
\begin{center}
\setlength{\unitlength}{0.008125in}%
\begingroup\makeatletter\ifx\SetFigFont\undefined
\def\x#1#2#3#4#5#6#7\relax{\def\x{#1#2#3#4#5#6}}%
\expandafter\x\fmtname xxxxxx\relax \def\y{splain}%
\ifx\x\y   
\gdef\SetFigFont#1#2#3{%
  \ifnum #1<17\tiny\else \ifnum #1<20\small\else
  \ifnum #1<24\normalsize\else \ifnum #1<29\large\else
  \ifnum #1<34\Large\else \ifnum #1<41\LARGE\else
     \huge\fi\fi\fi\fi\fi\fi
  \csname #3\endcsname}%
\else
\gdef\SetFigFont#1#2#3{\begingroup
  \count@#1\relax \ifnum 25<\count@\count@25\fi
  \def\x{\endgroup\@setsize\SetFigFont{#2pt}}%
  \expandafter\x
    \csname \romannumeral\the\count@ pt\expandafter\endcsname
    \csname @\romannumeral\the\count@ pt\endcsname
  \csname #3\endcsname}%
\fi
\fi\endgroup
\begin{picture}(290,280)(95,505)
\thinlines
\put(225,695){\line(-1, 0){  5}}
\put(220,695){\line( 0,-1){ 35}}
\put(220,660){\line( 1, 0){  5}}
\put(260,695){\line( 1, 0){  5}}
\put(265,695){\line( 0,-1){ 35}}
\put(265,660){\line(-1, 0){  5}}
\put(230,665){\makebox(0,0)[lb]{\smash{\SetFigFont{7}{8.4}{rm}fin:}}}
\put(230,685){\makebox(0,0)[lb]{\smash{\SetFigFont{7}{8.4}{rm}top:}}}
\put(230,675){\makebox(0,0)[lb]{\smash{\SetFigFont{7}{8.4}{rm}wh:}}}
\put(255,675){\makebox(0,0)[lb]{\smash{\SetFigFont{7}{8.4}{rm}-}}}
\put(255,665){\makebox(0,0)[lb]{\smash{\SetFigFont{7}{8.4}{rm}+}}}
\put(255,685){\makebox(0,0)[lb]{\smash{\SetFigFont{7}{8.4}{rm}-}}}
\put(345,570){\line(-1, 0){  5}}
\put(340,570){\line( 0,-1){ 35}}
\put(340,535){\line( 1, 0){  5}}
\put(380,570){\line( 1, 0){  5}}
\put(385,570){\line( 0,-1){ 35}}
\put(385,535){\line(-1, 0){  5}}
\put(350,540){\makebox(0,0)[lb]{\smash{\SetFigFont{7}{8.4}{rm}fin:}}}
\put(350,560){\makebox(0,0)[lb]{\smash{\SetFigFont{7}{8.4}{rm}top:}}}
\put(350,550){\makebox(0,0)[lb]{\smash{\SetFigFont{7}{8.4}{rm}wh:}}}
\put(375,550){\makebox(0,0)[lb]{\smash{\SetFigFont{7}{8.4}{rm}-}}}
\put(375,540){\makebox(0,0)[lb]{\smash{\SetFigFont{7}{8.4}{rm}+}}}
\put(375,560){\makebox(0,0)[lb]{\smash{\SetFigFont{7}{8.4}{rm}-}}}
\put(205,785){\line(-1, 0){  5}}
\put(200,785){\line( 0,-1){ 35}}
\put(200,750){\line( 1, 0){  5}}
\put(240,785){\line( 1, 0){  5}}
\put(245,785){\line( 0,-1){ 35}}
\put(245,750){\line(-1, 0){  5}}
\put(210,755){\makebox(0,0)[lb]{\smash{\SetFigFont{7}{8.4}{rm}fin:}}}
\put(210,775){\makebox(0,0)[lb]{\smash{\SetFigFont{7}{8.4}{rm}top:}}}
\put(210,765){\makebox(0,0)[lb]{\smash{\SetFigFont{7}{8.4}{rm}wh:}}}
\put(235,755){\makebox(0,0)[lb]{\smash{\SetFigFont{7}{8.4}{rm}+}}}
\put(235,775){\makebox(0,0)[lb]{\smash{\SetFigFont{7}{8.4}{rm}+}}}
\put(235,765){\makebox(0,0)[lb]{\smash{\SetFigFont{7}{8.4}{rm}-}}}
\put(100,725){\line(-1, 0){  5}}
\put( 95,725){\line( 0,-1){ 25}}
\put( 95,700){\line( 1, 0){  5}}
\put(135,725){\line( 1, 0){  5}}
\put(140,725){\line( 0,-1){ 25}}
\put(140,700){\line(-1, 0){  5}}
\put(105,715){\makebox(0,0)[lb]{\smash{\SetFigFont{7}{8.4}{rm}top:}}}
\put(105,705){\makebox(0,0)[lb]{\smash{\SetFigFont{7}{8.4}{rm}wh:}}}
\put(130,715){\makebox(0,0)[lb]{\smash{\SetFigFont{7}{8.4}{rm}+}}}
\put(130,705){\makebox(0,0)[lb]{\smash{\SetFigFont{7}{8.4}{rm}-}}}
\multiput(200,700)(0.00000,-8.00000){3}{\line( 0,-1){  4.000}}
\put(180,760){\line( 1,-2){ 20}}
\put(180,760){\line(-1,-2){ 20}}
\put(160,640){\line( 2, 1){ 40}}
\put(200,660){\line( 2,-1){ 40}}
\put(200,600){\line( 2, 1){ 40}}
\put(240,620){\line( 2,-1){ 40}}
\put(240,560){\line( 2, 1){ 40}}
\put(280,580){\line( 2,-1){ 40}}
\put(240,540){\line( 0,-1){ 20}}
\put(200,580){\line( 0,-1){ 20}}
\put(180,765){\makebox(0,0)[b]{\smash{\SetFigFont{7}{8.4}{rm}VP}}}
\put(200,705){\makebox(0,0)[b]{\smash{\SetFigFont{7}{8.4}{rm}VP}}}
\put(160,705){\makebox(0,0)[b]{\smash{\SetFigFont{7}{8.4}{rm}NP}}}
\put(200,665){\makebox(0,0)[b]{\smash{\SetFigFont{7}{8.4}{rm}VP}}}
\put(240,625){\makebox(0,0)[b]{\smash{\SetFigFont{7}{8.4}{rm}VP}}}
\put(160,625){\makebox(0,0)[b]{\smash{\SetFigFont{7}{8.4}{rm}Aux}}}
\put(200,585){\makebox(0,0)[b]{\smash{\SetFigFont{7}{8.4}{rm}NP}}}
\put(280,585){\makebox(0,0)[b]{\smash{\SetFigFont{7}{8.4}{rm}VP}}}
\put(240,545){\makebox(0,0)[b]{\smash{\SetFigFont{7}{8.4}{rm}V}}}
\put(200,545){\makebox(0,0)[b]{\smash{\SetFigFont{7}{8.4}{rm}e}}}
\put(240,505){\makebox(0,0)[b]{\smash{\SetFigFont{7}{8.4}{rm}baasaan}}}
\put(320,545){\makebox(0,0)[b]{\smash{\SetFigFont{7}{8.4}{rm}VP}}}
\put(160,685){\makebox(0,0)[b]{\smash{\SetFigFont{7}{8.4}{rm}(rameshas)}}}
\put(160,605){\makebox(0,0)[b]{\smash{\SetFigFont{7}{8.4}{rm}(chu)}}}
\end{picture}

\bigskip

\setlength{\unitlength}{0.008125in}%
\begingroup\makeatletter\ifx\SetFigFont\undefined
\def\x#1#2#3#4#5#6#7\relax{\def\x{#1#2#3#4#5#6}}%
\expandafter\x\fmtname xxxxxx\relax \def\y{splain}%
\ifx\x\y   
\gdef\SetFigFont#1#2#3{%
  \ifnum #1<17\tiny\else \ifnum #1<20\small\else
  \ifnum #1<24\normalsize\else \ifnum #1<29\large\else
  \ifnum #1<34\Large\else \ifnum #1<41\LARGE\else
     \huge\fi\fi\fi\fi\fi\fi
  \csname #3\endcsname}%
\else
\gdef\SetFigFont#1#2#3{\begingroup
  \count@#1\relax \ifnum 25<\count@\count@25\fi
  \def\x{\endgroup\@setsize\SetFigFont{#2pt}}%
  \expandafter\x
    \csname \romannumeral\the\count@ pt\expandafter\endcsname
    \csname @\romannumeral\the\count@ pt\endcsname
  \csname #3\endcsname}%
\fi
\fi\endgroup
\begin{picture}(227,315)(100,465)
\thinlines
\put(205,780){\line(-1, 0){  5}}
\put(200,780){\line( 0,-1){ 35}}
\put(200,745){\line( 1, 0){  5}}
\put(240,780){\line( 1, 0){  5}}
\put(245,780){\line( 0,-1){ 35}}
\put(245,745){\line(-1, 0){  5}}
\put(210,750){\makebox(0,0)[lb]{\smash{\SetFigFont{7}{8.4}{rm}fin:}}}
\put(210,770){\makebox(0,0)[lb]{\smash{\SetFigFont{7}{8.4}{rm}top:}}}
\put(210,760){\makebox(0,0)[lb]{\smash{\SetFigFont{7}{8.4}{rm}wh:}}}
\put(235,750){\makebox(0,0)[lb]{\smash{\SetFigFont{7}{8.4}{rm}+}}}
\put(235,760){\makebox(0,0)[lb]{\smash{\SetFigFont{7}{8.4}{rm}+}}}
\put(225,695){\line(-1, 0){  5}}
\put(220,695){\line( 0,-1){ 35}}
\put(220,660){\line( 1, 0){  5}}
\put(260,695){\line( 1, 0){  5}}
\put(265,695){\line( 0,-1){ 35}}
\put(265,660){\line(-1, 0){  5}}
\put(230,665){\makebox(0,0)[lb]{\smash{\SetFigFont{7}{8.4}{rm}fin:}}}
\put(230,685){\makebox(0,0)[lb]{\smash{\SetFigFont{7}{8.4}{rm}top:}}}
\put(230,675){\makebox(0,0)[lb]{\smash{\SetFigFont{7}{8.4}{rm}wh:}}}
\put(255,675){\makebox(0,0)[lb]{\smash{\SetFigFont{7}{8.4}{rm}-}}}
\put(255,665){\makebox(0,0)[lb]{\smash{\SetFigFont{7}{8.4}{rm}+}}}
\multiput(200,700)(0.00000,-8.00000){3}{\line( 0,-1){  4.000}}
\put(160,640){\line( 2, 1){ 40}}
\put(200,660){\line( 2,-1){ 40}}
\put(200,600){\line( 2, 1){ 40}}
\put(240,620){\line( 2,-1){ 40}}
\put(240,560){\line( 2, 1){ 40}}
\put(280,580){\line( 2,-1){ 40}}
\put(240,540){\line( 0,-1){ 20}}
\put(320,540){\line( 0,-1){ 20}}
\put(320,500){\line( 0,-1){ 20}}
\put(200,705){\makebox(0,0)[b]{\smash{\SetFigFont{7}{8.4}{rm}VP}}}
\put(240,625){\makebox(0,0)[b]{\smash{\SetFigFont{7}{8.4}{rm}VP}}}
\put(200,585){\makebox(0,0)[b]{\smash{\SetFigFont{7}{8.4}{rm}NP}}}
\put(280,585){\makebox(0,0)[b]{\smash{\SetFigFont{7}{8.4}{rm}VP}}}
\put(240,545){\makebox(0,0)[b]{\smash{\SetFigFont{7}{8.4}{rm}NP}}}
\put(320,545){\makebox(0,0)[b]{\smash{\SetFigFont{7}{8.4}{rm}VP}}}
\put(320,465){\makebox(0,0)[b]{\smash{\SetFigFont{7}{8.4}{rm}kor}}}
\put(320,505){\makebox(0,0)[b]{\smash{\SetFigFont{7}{8.4}{rm}V}}}
\put(240,505){\makebox(0,0)[b]{\smash{\SetFigFont{7}{8.4}{rm}e}}}
\put(200,665){\makebox(0,0)[b]{\smash{\SetFigFont{7}{8.4}{rm}VP}}}
\put(160,625){\makebox(0,0)[b]{\smash{\SetFigFont{7}{8.4}{rm}COMP}}}
\put(140,725){\line( 1, 0){  5}}
\put(145,725){\line( 0,-1){ 25}}
\put(145,700){\line(-1, 0){  5}}
\put(105,725){\line(-1, 0){  5}}
\put(100,725){\line( 0,-1){ 25}}
\put(100,700){\line( 1, 0){  5}}
\put(110,715){\makebox(0,0)[lb]{\smash{\SetFigFont{7}{8.4}{rm}top:}}}
\put(110,705){\makebox(0,0)[lb]{\smash{\SetFigFont{7}{8.4}{rm}wh:}}}
\put(135,705){\makebox(0,0)[lb]{\smash{\SetFigFont{7}{8.4}{rm}+}}}
\put(180,760){\line( 1,-2){ 20}}
\put(180,760){\line(-1,-2){ 20}}
\put(180,765){\makebox(0,0)[b]{\smash{\SetFigFont{7}{8.4}{rm}VP}}}
\put(160,705){\makebox(0,0)[b]{\smash{\SetFigFont{7}{8.4}{rm}NP}}}
\put(200,565){\makebox(0,0)[b]{\smash{\SetFigFont{7}{8.4}{rm}(me)}}}
\put(160,605){\makebox(0,0)[b]{\smash{\SetFigFont{7}{8.4}{rm}(ki)}}}
\put(160,685){\makebox(0,0)[b]{\smash{\SetFigFont{7}{8.4}{rm}(kyaa)}}}
\end{picture}
\end{center}
\caption{D-trees for (\protect\ref{sent-kash}b)}
\label{fig-kash-matrix}
\end{figure}

Figure~\ref{fig-kash-matrix} shows the matrix and embedded clauses for
sentence (\ref{sent-kash}b).  We use the node label VP throughout and use
features such as {\featf top} (for topic) to differentiate different
levels of projection.  Observe that in both trees an argument has been
fronted.  Again, we will use the {\SIC}s to enforce the projection from a
lexical anchor to its maximal projection.  Since the direct object of
{\em kor} has {\em wh}-moved out of its clause, the d-edge connecting
it to the maximal projection of its verb has no {\SIC}\@.  The d-edge
connecting the maximal projection of {\em baasaan} to the Aux component,
however, has a {\SIC} that allows only  VP[{\featf wh}: +, {\featf top}:
-] nodes to be inserted.

\begin{figure}[htb]
\begin{center}
\setlength{\unitlength}{0.008125in}%
\begingroup\makeatletter\ifx\SetFigFont\undefined
\def\x#1#2#3#4#5#6#7\relax{\def\x{#1#2#3#4#5#6}}%
\expandafter\x\fmtname xxxxxx\relax \def\y{splain}%
\ifx\x\y   
\gdef\SetFigFont#1#2#3{%
  \ifnum #1<17\tiny\else \ifnum #1<20\small\else
  \ifnum #1<24\normalsize\else \ifnum #1<29\large\else
  \ifnum #1<34\Large\else \ifnum #1<41\LARGE\else
     \huge\fi\fi\fi\fi\fi\fi
  \csname #3\endcsname}%
\else
\gdef\SetFigFont#1#2#3{\begingroup
  \count@#1\relax \ifnum 25<\count@\count@25\fi
  \def\x{\endgroup\@setsize\SetFigFont{#2pt}}%
  \expandafter\x
    \csname \romannumeral\the\count@ pt\expandafter\endcsname
    \csname @\romannumeral\the\count@ pt\endcsname
  \csname #3\endcsname}%
\fi
\fi\endgroup
\begin{picture}(326,500)(61,285)
\thinlines
\put(105,740){\line(-1, 0){  5}}
\put(100,740){\line( 0,-1){ 25}}
\put(100,715){\line( 1, 0){  5}}
\put(140,740){\line( 1, 0){  5}}
\put(145,740){\line( 0,-1){ 25}}
\put(145,715){\line(-1, 0){  5}}
\put(110,730){\makebox(0,0)[lb]{\smash{\SetFigFont{7}{8.4}{rm}top:}}}
\put(110,720){\makebox(0,0)[lb]{\smash{\SetFigFont{7}{8.4}{rm}wh:}}}
\put(135,730){\makebox(0,0)[lb]{\smash{\SetFigFont{7}{8.4}{rm}+}}}
\put(135,720){\makebox(0,0)[lb]{\smash{\SetFigFont{7}{8.4}{rm}-}}}
\put(140,680){\line(-1, 0){  5}}
\put(135,680){\line( 0,-1){ 25}}
\put(135,655){\line( 1, 0){  5}}
\put(175,680){\line( 1, 0){  5}}
\put(180,680){\line( 0,-1){ 25}}
\put(180,655){\line(-1, 0){  5}}
\put(145,670){\makebox(0,0)[lb]{\smash{\SetFigFont{7}{8.4}{rm}top:}}}
\put(145,660){\makebox(0,0)[lb]{\smash{\SetFigFont{7}{8.4}{rm}wh:}}}
\put(170,660){\makebox(0,0)[lb]{\smash{\SetFigFont{7}{8.4}{rm}+}}}
\put(160,785){\line(-1, 0){  5}}
\put(155,785){\line( 0,-1){ 35}}
\put(155,750){\line( 1, 0){  5}}
\put(195,785){\line( 1, 0){  5}}
\put(200,785){\line( 0,-1){ 35}}
\put(200,750){\line(-1, 0){  5}}
\put(165,755){\makebox(0,0)[lb]{\smash{\SetFigFont{7}{8.4}{rm}fin:}}}
\put(165,775){\makebox(0,0)[lb]{\smash{\SetFigFont{7}{8.4}{rm}top:}}}
\put(165,765){\makebox(0,0)[lb]{\smash{\SetFigFont{7}{8.4}{rm}wh:}}}
\put(190,755){\makebox(0,0)[lb]{\smash{\SetFigFont{7}{8.4}{rm}+}}}
\put(190,775){\makebox(0,0)[lb]{\smash{\SetFigFont{7}{8.4}{rm}+}}}
\put(190,765){\makebox(0,0)[lb]{\smash{\SetFigFont{7}{8.4}{rm}-}}}
\put(195,725){\line(-1, 0){  5}}
\put(190,725){\line( 0,-1){ 35}}
\put(190,690){\line( 1, 0){  5}}
\put(230,725){\line( 1, 0){  5}}
\put(235,725){\line( 0,-1){ 35}}
\put(235,690){\line(-1, 0){  5}}
\put(200,695){\makebox(0,0)[lb]{\smash{\SetFigFont{7}{8.4}{rm}fin:}}}
\put(200,715){\makebox(0,0)[lb]{\smash{\SetFigFont{7}{8.4}{rm}top:}}}
\put(200,705){\makebox(0,0)[lb]{\smash{\SetFigFont{7}{8.4}{rm}wh:}}}
\put(225,695){\makebox(0,0)[lb]{\smash{\SetFigFont{7}{8.4}{rm}+}}}
\put(225,705){\makebox(0,0)[lb]{\smash{\SetFigFont{7}{8.4}{rm}+}}}
\put(245,670){\line(-1, 0){  5}}
\put(240,670){\line( 0,-1){ 35}}
\put(240,635){\line( 1, 0){  5}}
\put(280,670){\line( 1, 0){  5}}
\put(285,670){\line( 0,-1){ 35}}
\put(285,635){\line(-1, 0){  5}}
\put(250,640){\makebox(0,0)[lb]{\smash{\SetFigFont{7}{8.4}{rm}fin:}}}
\put(250,660){\makebox(0,0)[lb]{\smash{\SetFigFont{7}{8.4}{rm}top:}}}
\put(250,650){\makebox(0,0)[lb]{\smash{\SetFigFont{7}{8.4}{rm}wh:}}}
\put(275,650){\makebox(0,0)[lb]{\smash{\SetFigFont{7}{8.4}{rm}-}}}
\put(275,640){\makebox(0,0)[lb]{\smash{\SetFigFont{7}{8.4}{rm}+}}}
\put(275,660){\makebox(0,0)[lb]{\smash{\SetFigFont{7}{8.4}{rm}-}}}
\put(220,500){\line( 0,-1){ 20}}
\put(220,505){\makebox(0,0)[b]{\smash{\SetFigFont{9}{10.8}{rm}V}}}
\put(305,520){\line(-1, 0){  5}}
\put(300,520){\line( 0,-1){ 35}}
\put(300,485){\line( 1, 0){  5}}
\put(340,520){\line( 1, 0){  5}}
\put(345,520){\line( 0,-1){ 35}}
\put(345,485){\line(-1, 0){  5}}
\put(310,490){\makebox(0,0)[lb]{\smash{\SetFigFont{7}{8.4}{rm}fin:}}}
\put(310,510){\makebox(0,0)[lb]{\smash{\SetFigFont{7}{8.4}{rm}top:}}}
\put(310,500){\makebox(0,0)[lb]{\smash{\SetFigFont{7}{8.4}{rm}wh:}}}
\put(335,500){\makebox(0,0)[lb]{\smash{\SetFigFont{7}{8.4}{rm}-}}}
\put(335,490){\makebox(0,0)[lb]{\smash{\SetFigFont{7}{8.4}{rm}+}}}
\put(335,510){\makebox(0,0)[lb]{\smash{\SetFigFont{7}{8.4}{rm}-}}}
\put(250,440){\line( 0,-1){ 20}}
\put(250,405){\makebox(0,0)[b]{\smash{\SetFigFont{9}{10.8}{rm}ki}}}
\put(210,560){\line( 0,-1){ 20}}
\put(210,565){\makebox(0,0)[b]{\smash{\SetFigFont{7}{8.4}{rm}NP}}}
\put(210,525){\makebox(0,0)[b]{\smash{\SetFigFont{7}{8.4}{rm}e}}}
\put( 80,720){\line( 0,-1){ 20}}
\put( 80,740){\line( 3, 1){ 60}}
\put(140,760){\line( 1,-2){ 20}}
\put(120,680){\line( 2, 1){ 40}}
\put(120,660){\line( 0,-1){ 20}}
\put(160,700){\line( 3,-2){ 60}}
\put(180,620){\line( 2, 1){ 40}}
\put(180,600){\line( 0,-1){ 20}}
\put(220,640){\line( 1,-1){ 20}}
\put(240,600){\line( 1,-2){ 20}}
\put(220,520){\line( 2, 1){ 40}}
\put(260,540){\line( 1,-2){ 20}}
\put(280,480){\line( 2,-1){ 40}}
\put(320,360){\line( 0,-1){ 20}}
\put(280,400){\line( 0,-1){ 20}}
\put(250,460){\line( 3, 2){ 30}}
\put(380,360){\line( 0,-1){ 20}}
\put(380,320){\line( 0,-1){ 20}}
\put(280,420){\line( 2, 1){ 40}}
\put(320,440){\line( 3,-2){ 30}}
\put(320,380){\line( 3, 2){ 30}}
\put(350,400){\line( 3,-2){ 30}}
\put(240,600){\line(-3,-2){ 30}}
\put(140,765){\makebox(0,0)[b]{\smash{\SetFigFont{7}{8.4}{rm}VP}}}
\put( 80,725){\makebox(0,0)[b]{\smash{\SetFigFont{7}{8.4}{rm}NP}}}
\put( 80,685){\makebox(0,0)[b]{\smash{\SetFigFont{7}{8.4}{rm}rameshas}}}
\put(120,665){\makebox(0,0)[b]{\smash{\SetFigFont{7}{8.4}{rm}NP}}}
\put(160,705){\makebox(0,0)[b]{\smash{\SetFigFont{7}{8.4}{rm}VP}}}
\put(120,625){\makebox(0,0)[b]{\smash{\SetFigFont{7}{8.4}{rm}kyaa}}}
\put(180,605){\makebox(0,0)[b]{\smash{\SetFigFont{7}{8.4}{rm}Aux}}}
\put(220,645){\makebox(0,0)[b]{\smash{\SetFigFont{7}{8.4}{rm}VP}}}
\put(180,565){\makebox(0,0)[b]{\smash{\SetFigFont{7}{8.4}{rm}chu}}}
\put(240,605){\makebox(0,0)[b]{\smash{\SetFigFont{7}{8.4}{rm}VP}}}
\put(260,545){\makebox(0,0)[b]{\smash{\SetFigFont{7}{8.4}{rm}VP}}}
\put(220,465){\makebox(0,0)[b]{\smash{\SetFigFont{7}{8.4}{rm}baasaan}}}
\put(320,445){\makebox(0,0)[b]{\smash{\SetFigFont{9}{10.8}{rm}VP}}}
\put(320,325){\makebox(0,0)[b]{\smash{\SetFigFont{9}{10.8}{rm}e}}}
\put(280,485){\makebox(0,0)[b]{\smash{\SetFigFont{7}{8.4}{rm}VP}}}
\put(280,405){\makebox(0,0)[b]{\smash{\SetFigFont{7}{8.4}{rm}NP}}}
\put(320,365){\makebox(0,0)[b]{\smash{\SetFigFont{7}{8.4}{rm}NP}}}
\put(380,365){\makebox(0,0)[b]{\smash{\SetFigFont{7}{8.4}{rm}VP}}}
\put(380,325){\makebox(0,0)[b]{\smash{\SetFigFont{7}{8.4}{rm}V}}}
\put(380,285){\makebox(0,0)[b]{\smash{\SetFigFont{7}{8.4}{rm}kor}}}
\put(280,365){\makebox(0,0)[b]{\smash{\SetFigFont{7}{8.4}{rm}me}}}
\put(350,405){\makebox(0,0)[b]{\smash{\SetFigFont{7}{8.4}{rm}VP}}}
\put(250,445){\makebox(0,0)[b]{\smash{\SetFigFont{7}{8.4}{rm}COMP}}}
\end{picture}
\end{center}
\caption{Derived d-tree for (\protect\ref{sent-kash}b)}
\label{fig-kash-derived}
\end{figure}
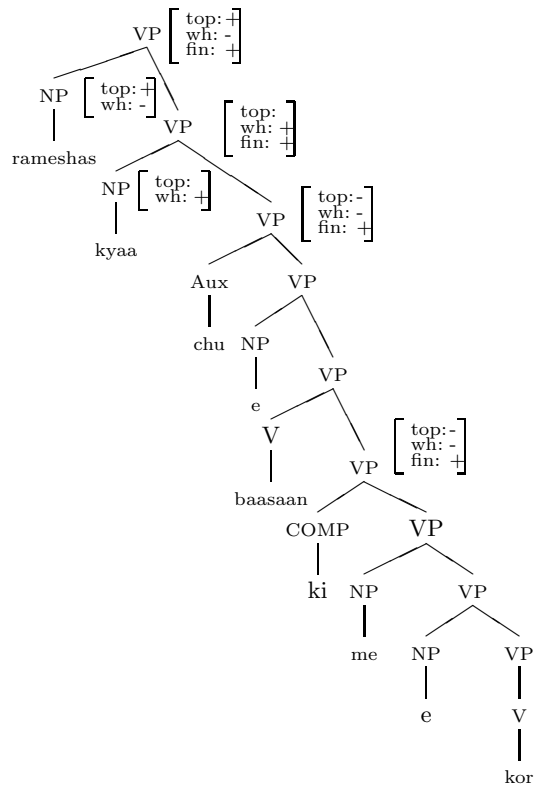

The derivation proceeds as follows.  We first subsert the embedded
clause tree into the matrix clause tree.  After that, we subsert the
nominal arguments and function words.  The derived structure is shown in
Figure~\ref{fig-kash-derived}.  The associated {\SAtree} is the desired,
semantically motivated, dependency structure: the embedded clause
depends on the matrix clause.

In this section, we have discussed examples where the elementary objects
have been obtained by projecting from lexical items. In these cases, we
overcome both the problems with {\TAG} considered in
Section~\ref{sec-intro}.  The {\SIC}s considered here enforce the same
notion of projection that was used in obtaining the elementary
structures. This method of arriving at {\SIC}s not only generalizes for
the English and Kashmiri examples but also appears to apply to the case
of long-distance scrambling and topicalization in German.

\section{Recognition}
\label{sec-properties}

It is straightforward to adapt the polynomial-time {\CKY}-style recognition
algorithm for a lexicalized {\UVGDL} of Rambow~\shortcite{rambow:94c} for
{\DTG}\@.  The entries in this array recording derivations of substrings of
input contain a set of elementary nodes along with a multi-set of
components that must be inserted above during bottom-up
recognition. These components are added or removed at substitution and
insertion. The algorithm simulates traversal of a derived tree; checking
for {\SIC}s and {\SAC}s can be done easily. Because of lexicalization, the
size of these multi-sets is polynomially bounded, from which the
polynomial time and space complexity of the algorithm follows.

For practical purposes, especially for lexicalized grammars, it is
preferable to incorporate some element of prediction. We are developing
a polynomial-time Earley style parsing algorithm. The parser returns a
parse forest encoding all parses for an input string. The performance
of this parser is sensitive to the grammar and input. Indeed it appears
that for grammars that lexicalize {\CFG} and for English grammar (where
the structures are similar to the {\LTAG} developed at University of
Pennsylvania~\cite{xtag95}) we obtain cubic-time complexity.

\section{Conclusion}

{\DTG}, like other formalisms in the {\TAG} family, is lexicalizable, but
in addition, its derivations are themselves linguistically meaningful.
In future work we intend to examine additional linguistic data, refining
aspects of our definition as needed.  We will also study the formal
properties of {\DTG}, and complete the design of the Earley style parser.

\section*{Acknowledgements}
We would like to thank Rakesh Bhatt for help with the Kashmiri data.  We
are also grateful to Tilman Becker, Gerald Gazdar, Aravind Joshi, Bob
Kasper, Bill Keller, Tony Kroch, Klaus Netter and the ACL-95 referees.
Rambow was supported by the North Atlantic Treaty Organization under a
Grant awarded in 1993, while at TALANA,
Universit{\'e} Paris 7.

\end{document}